\documentclass{article}
\usepackage[final]{corl_2023}

\usepackage[hidelinks]{hyperref}
\usepackage{cite}
\usepackage[linesnumbered,ruled,vlined]{algorithm2e}
\usepackage{adjustbox}

\usepackage{mathrsfs}

\usepackage{blindtext}
\usepackage{multicol}
\usepackage{subcaption}
\usepackage{caption}
\usepackage{color,soul}
\usepackage[T1]{fontenc}
\usepackage[table,xcdraw,dvipsnames]{xcolor}
\usepackage{xcolor}
\usepackage{textcomp}
\usepackage{graphicx}
\usepackage{algorithmic}
\usepackage{amsmath,amssymb,amsfonts}
\usepackage{float}
\usepackage{multirow}
\usepackage{booktabs,makecell}

\graphicspath{{figures/}} 

\begin{document}


\title{A Hierarchical Approach to Imitation Learning for Manipulation Tasks Requiring Time Varying Forces}

%
%


\author{
    Rishabh Shukla, Adithya Santhosh, Shaili Gandhi, Samrudh Moode, and Satyandra K. Gupta \\
    Realization of Robotic Systems Lab (RROS) \\
    University of Southern California, Los Angeles, CA, USA 
    \thanks{Address all correspondence to {\tt\small \href{mailto:guptask@usc.edu}{guptask@usc.edu}}}
    }




\maketitle
\begin{abstract} 

Diffusion policies have shown strong performance in learning complex, multi-modal behaviors for robotic manipulation. However, their application to contact-rich disassembly tasks remains limited by a key trade-off: the iterative denoising process introduces inference latencies that makes high frequency control difficult, which is essential for realizing dynamic interactions such as chiseling and prying. Recent action-chunking techniques mitigate latency but use an open-loop execution window, rendering the system blind to rapid force transients caused by fracture events. To bridge this gap, we introduce the Diffusion Policy Augmented by Fast Trajectory Generation (DPA-FTG). Compared to recent visual-tactile approaches that focus on positional correction, DPA-FTG decouples low-frequency planning from high-frequency force regulation. At the high level ($5$ Hz), a conditional diffusion model predicts a sequence of latent parameters for selecting a strategy from a learned vocabulary of task primitives. At the low level ($60$ Hz), a lightweight, force-conditioned policy acts as a neural impedance controller, modulating execution in real-time to maintain contact stability. We validate our approach on a bimanual battery disassembly task involving the separation of a compliant sheet. Experimental evaluation demonstrates that DPA-FTG outperforms state-of-the-art baselines, including Reactive Diffusion Policy (RDP).

\end{abstract}

\section{Introduction}
\begin{figure*}[h]
    \centering
    \includegraphics[width=\linewidth]{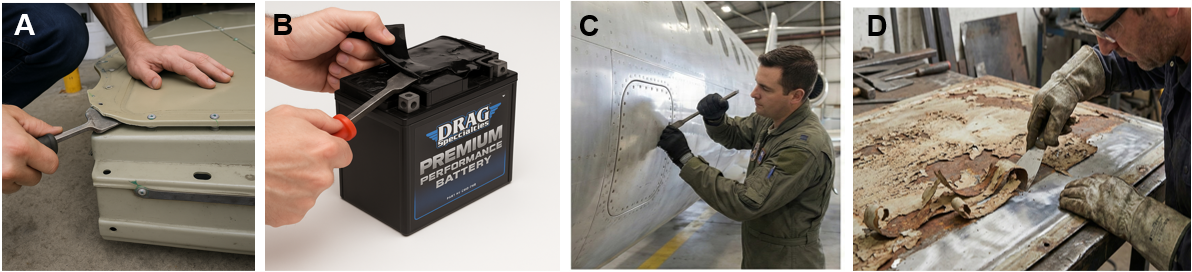}
     \caption{Material separation tasks require complex force modulation across diverse domains. (A) Separating compliant busbars in EV battery remanufacturing. (B) Manual prying of bonded components. (C) Removing sealant or panels in aerospace maintenance. (D) Scraping corrosion or fused layers in general salvage operations. While visually distinct, these tasks share a common dynamic profile: high-frequency force application coupled with low-frequency trajectory planning.}
    \label{fig:task_generalization}   
\end{figure*}

\noindent Material separation and surface preparation are critical tasks in industrial automation. Processes such as chiseling, scraping, peeling, and prying are essential for operations ranging from aerospace maintenance (e.g., sealant removal) to infrastructure repair (e.g., removing corrosion) and component salvage. These tasks are currently performed manually due to the complexity of the required tool manipulation. As illustrated in Fig. \ref{fig:task_generalization}, these operations require bimanual coordination to simultaneously break adhesion (chisel) and manage material deformation (peel). Whether removing sealant from an aircraft service hatch (Fig. \ref{fig:task_generalization}C), stripping rusted panels (Fig. \ref{fig:task_generalization}D), or disassembling Electric Vehicle (EV) batteries (Fig. \ref{fig:task_generalization}A, \ref{fig:task_generalization}B), the underlying tool-part interaction requirements are similar. In each scenario, the system must exert precise, often oscillatory forces to fracture bonds without damaging the substrate or the tool.

Unlike structured assembly tasks where components are rigid, geometries are known, and poses are deterministic, separation tasks often occur in  semi-structured environments. These scenarios are characterized by deformable materials, unknown surface characteristics, and, most critically, unpredictable bond strengths. A canonical challenge in this domain is the separation of compliant sheets adhered to rigid substrates. This requires a dual-arm system to effectively decouple the dynamics of adhesion breaking from material handling; one arm must stabilize or peel the component while the other modulates the impedance of a tool to fracture the bond \cite{shukla_force-conditioned_nodate}. Furthermore, these interactions are not static. Breaking adhesive bonds often excites high-frequency dynamics, requiring oscillatory tool motions exceeding 10 Hz to initiate fracture without damaging the substrate. Consequently, modulating these rapid contact forces necessitates a closed-loop control frequency (e.g., $60$ Hz) significantly higher than standard manipulation planning loops.

Current automation techniques, particularly those relying on traditional position control, often struggle to scale to these high-mix, contact-rich scenarios. The rigidity of standard industrial automation is ill-suited for the multi-modal nature of these tasks. This is because the optimal policy is not a single deterministic trajectory, but rather a set of distinct valid strategies or task primitives (e.g., varying tool attack angles or switching between steady pushing and oscillatory prying) that must be selected based on local conditions. In such a challenging domain, a millimeter of position error can result in either a missed cut or damage to the workpiece. This limitation is particularly acute given the growing shortage of skilled labor willing to undertake these repetitive, physically demanding, and hazardous disassembly tasks \cite{ev_recycle, hellmuth2021assessment}. Therefore, there is a need for robotic systems that can learn these dexterous skills from demonstration and generalize them across varying geometries and material properties. However, developing an analytical controller for such tasks is challenging. 

The interactive dynamics between components for the separation task is governed by unknown friction coefficients, variable adhesive properties, and material deformation, and is often intractable to model explicitly using traditional model-based approaches. Consequently, the field has shifted towards data-driven methods that can implicitly capture these complex strategies directly from expert behavior. Imitation Learning (IL) has emerged as a promising avenue for acquiring such skills from human demonstration \cite{hussein2017imitation, fang2019survey}. Specifically, Diffusion Policies (DP) \cite{chi2023diffusion, chi2024universal, kang2025robotic} represent the current state-of-the-art, uniquely capable of capturing the multi-modal distributions of expert behavior discussed above and overcoming the limitations of traditional behavioral cloning \cite{florence2022implicit}.

However, standard DPs are inherently limited by their inference speed. Breaking adhesive bonds often requires oscillatory chiseling at frequencies exceeding 10 Hz, necessitating a control loop of at least 60 Hz to prevent tool slip or workpiece damage. Standard DPs, operating at lower frequencies (approx. 5–10 Hz), cannot react fast enough to sudden changes in bond dynamics (e.g., instantaneous fracture) \cite{xue_reactive_2025}. Another critical limitation in current State-of-the-Art DPs is that they typically rely on action chunking, i.e., predicting a sequence of actions and executing them open-loop. While excellent for kinematic trajectory generation, this open-loop execution renders the robot unresponsive to the rapid force feedback required during physical interaction. While standard methods utilizing Action Chunking, have successfully solved the problem of temporal consistency in free-space motion, they introduce a fatal flaw for material removal, i.e., the open-loop execution. In stiff contact scenarios, environment dynamics evolve faster than this execution window.  Even if inference speeds were accelerated (e.g., using Consistency Models), the system remains open-loop for a significant time window. In stiff contact tasks like chiseling, the duration of this open-loop commitment is sufficient to cause tool damage. Consequently, to solve this problem, we must decouple task selection from fast reactive execution.

In our exploratory work \cite{shukla_force-conditioned_nodate}, we evaluated the use of a diffusion-based visuomotor policy augmented with force feedback for compliant sheet separation. While this demonstrated that force sensing improves precision, the system remained constrained by the inference latency of the diffusion model. The policy operated at a modest frequency, forcing the robot to slow down significantly to maintain safety, thereby prolonging the task execution time. Recent advancements such as Reactive Diffusion Policy (RDP) \cite{xue_reactive_2025} have attempted to bridge this gap using "slow-fast" architectures. However, while RDP succeeds in positional correction based on tactile deformation, it does not explicitly model the dynamic force regulation required for chiseling. The goal in these tasks is not merely to reach a Cartesian pose, but to apply a specific force profile to detach components.

To address these limitations, we propose \textbf{Diffusion Policy Augmented by Fast Trajectory Generation (DPA-FTG)}. We formulate the problem as a hierarchical decomposition of Task Selection and Trajectory Generation. To do this, we shift the concept of 'chunking' from the Action Space to the Latent Space. We posit that, while the execution of a skill must be high-frequency and reactive, the intent of that skill remains stable over longer horizons. Instead of diffusing a rigid sequence of poses that blinds the robot to contact dynamics, our system predicts a Latent Primitive i.e. a compact encoding of the strategic intent (e.g., 'oscillate forward' versus 'pry upward').

This stable intent is passed to a high-frequency Force-Reactive Decoder. Unlike standard diffusion decoders that simply play back a fixed trajectory, our recurrent decoder unzips the latent intent into motor commands step-by-step at 60 Hz. This allows the system to modulate the trajectory inside the chunk. If a force spike is detected, the decoder can instantaneously deviate from the nominal path to prevent damage. This effectively closes the control loop during the inference window.

Thus, our architecture has two components:
\begin{enumerate}

\item Task Selection (High-Level): A diffusion model operating at 5 Hz observes the global state and visual context to perform task selection (low-frequency intent, e.g., "initiate chiseling at 15$^{\circ}$ pitch"). It outputs a latent command $z_k$.

\item Trajectory Generation (Low-Level Dynamics): A recurrent neural network operating at 60 Hz observes $z_k$ and high-frequency force/torque (F/T) data to modulate action. This layer allows for immediate reaction to force transients.

\end{enumerate}

To validate DPA-FTG, we conduct experiments on a generalized material separation task involving the removal of compliant adhesive sheets from rigid workpieces, as shown in Fig. \ref{fig:comparison_rigid_vs_deformable_swing}. While this setup serves as a surrogate for EV battery disassembly, it is designed to test the system across diverse sheet geometries and varying bond characteristics. This allows us to benchmark the approach against the core complexities without relying on specific production parts. DPA-FTG outperforms both standard DPs and reactive baselines in these contact-rich scenarios. Please checkout the video of our approach at the following link: \url{https://youtu.be/5pP2cpnkbrE}.

In summary, our contributions are as follows:

\begin{enumerate}

\item \textbf{DPA-FTG Architecture:} An integration of diffusion-based planning with high-frequency learned force control, enabling stable 60 Hz interaction for material separation tasks.

\item \textbf{Latent Primitive Learning:} A method for learning a compact latent space of oscillatory motion primitives that bridges the frequency gap between the visual planner and the force controller.

\item \textbf{Empirical Validation:} Benchmarking against RDP and standard DP on a bimanual robotic setup, improving success rates and speed in compliant sheet separation.

\end{enumerate}
\section{Related Work}
\label{sec:Related Work}
\noindent Diffusion models, originally developed for image generation \cite{ho_denoising_2020}, have recently gained traction in robotics for policy learning. They offer advantages over traditional methods like Behavioral Cloning (BC) \cite{mandlekar2021matters} and Generative Adversarial Networks (GANs) due to their ability to model complex, multimodal action distributions and their training stability \cite{wolf_diffusion_2025}. Chi et al. \cite{chi2023diffusion} introduced the Diffusion Policy, establishing a strong baseline for visuomotor control. Subsequent works have extended DPs to incorporate 3D representations (DP3) \cite{ze_3d_2024}, integrate language instructions \cite{wen_diffusionvla_2025}, and handle bimanual manipulation \cite{deng2025safebimanual}. Recent efforts have also focused on improving the efficiency and capability of these models \cite{batra2025zero}. Architectures like the Mamba Policy \cite{cao2025mamba} aim to reduce the computational overhead associated with standard DP backbones. Furthermore, recent surveys highlight the ongoing challenges and strategies for effectively incorporating multimodal feedback, particularly force and tactile data, into imitation learning for contact-rich tasks \cite{ito2025survey, li2025robotic}. Our prior work \cite{shukla_force-conditioned_nodate}, demonstrated the benefit of integrating force feedback into DPs for compliant separation tasks; however, that approach was still limited by the low control frequency inherent to standard DP architectures \cite{mishra2024generative, ren2024diffusion, janner2022planning}.


\textbf{Addressing Latency in Diffusion Policies.}
The primary bottleneck for deploying DPs in real-time control is the latency induced by the iterative denoising process. Research to address this fall into two main categories: algorithmic acceleration and improvements in architectural paradigms. Algorithmic Acceleration-based methods aim to speed up the generative core. Techniques like Denoising Diffusion Implicit Models (DDIMs) \cite{song_denoising_2022} optimize the sampling process. More recently, Consistency Models \cite{song_consistency_nodate, yan2025maniflow, lu2024manicm} have enabled one-step generation by learning a function that directly maps noise to data. The Consistency Policy \cite{prasad_consistency_2024} adapted this for robotics, achieving significant speedups.  This approach is utilized in recent real-world RL frameworks (e.g., RL-100 \cite{li2025rl100}) to distill multi-step diffusion into high-frequency one-step controllers. Efforts towards deployment efficiency, like LightDP \cite{wu2025ondevice}, combine network compression and consistency distillation to enable real-time execution on edge devices. However, distillation-based approaches may lead to a reduction in sample diversity or performance degradation compared to the full multi-step process \cite{zhang_real-time_2025}. 

Beyond distillation, other methods optimize the sampling strategy without retraining. Real Time Iteration-Diffusion Policy (RTI-DP) \cite{duan2025realtime} leverages the spatiotemporal consistency of physical systems by using the previous action sequence as an initialization for the next denoising process, accelerating inference. Genetic Denoising (GDP) \cite{clemente2025twosteps} further reduces the required steps to as few as two by employing a population-based sampling strategy tailored to the low-dimensional nature of action distributions. Other approaches include optimizing the training objective, such as Time Unified Diffusion Policy (TUDP) \cite{niu_time-unified_2025}. 

\textbf{Architectural methods that restructure the control loop.} Latent Weight Diffusion (LWD) \cite{hegde_latent_2025} proposes a paradigm shift by generating the weights of a small, reactive policy rather than trajectories. Rapid Adaptive-Diffusion Policy (RA-DP) \cite{ye_ra-dp_2025}  introduced an action queue mechanism for high-frequency replanning within the denoising loop. Beyond consistency models, recent work has explored alternative generative formulations to bypass iterative denoising entirely. The Energy Policy proposes learning a multimodal energy function that allows for single-pass action generation, demonstrating inference speeds 2–7 times faster than standard diffusion policies while maintaining multimodality \cite{jia2025fast}. Similarly, recent efforts in distilling diffusion models for high-frequency control have shown that while distillation can reduce latency, it often sacrifices the 'mode-covering' behavior critical for recovery in contact-rich tasks \cite{zhang2025real}.

\textbf{Hierarchical Control.} Separating high-level planning from low-level execution, is a well-established concept in robotics \cite{smith2012dual}. Modern learning-based approaches have adopted this "slow-fast" paradigm to manage computational complexity and improve reactivity \cite{stepputtis2022system, chen2022towards}. Hierarchical Diffusion Policy (HDP) \cite{ma_hierarchical_2024} factorizes manipulation into high-level goal prediction and low-level trajectory generation via a goal-conditioned diffusion policy. Hierarchical Robot Transform (HiRT) \cite{zhang_hirt_2025} integrates a slow Vision-Language Model (VLM) for planning with a fast execution module.

Highly relevant to our work is the Reactive Diffusion Policy (RDP) \cite{xue_reactive_2025}, which employs a slow latent diffusion policy and a fast asymmetric tokenizer for closed-loop tactile feedback control. RDP demonstrates the effectiveness of the slow-fast hierarchy for contact-rich manipulation. Our approach, DPA-FTG, builds upon this concept but differs in its objective. While RDP uses the fast policy primarily for reactive correction of the latent trajectory, DPA-FTG utilizes the fast policy to generate the high-frequency motion (e.g., oscillation) parameterized by the slow policy, enabling stable execution at higher frequencies (60 Hz) required for dynamic tasks like chiseling.

\textbf{High-Frequency and Force-Based Manipulation.} Contact-rich manipulation tasks often require high-frequency control and precise force regulation. Beyond diffusion-based imitation learning, these challenges are historically addressed using classical, non-generative algorithms. Frameworks such as impedance control \cite{hogan_impedance_1984} and hybrid force/motion control \cite{raibert_hybrid_1981} provide traditional structures for managing contact interactions. Similarly, template-based oscillation generators like Dynamic Movement Primitives (DMPs) \cite{ijspeert2013dynamical}, coupled with perception-driven keypoint generators, can produce specialized high-frequency behaviors. These non-Diffusion Policy approaches are effective when the task structure, contact models, or switching logic are known a priori. However, they typically require substantial manual design and cannot directly capture the multimodal strategy distributions present in human demonstrations. Consequently, while we include Behavioral Cloning (BC) (Sec.~\ref{subsec: baselines}) to evaluate non-diffusion alternatives, our primary experimental comparison focuses on learning-based baselines that share similar observation inputs and generalization requirements.

Within learning-based methods, policies have increasingly incorporated force feedback \cite{lin2023bi, helmut2025tactile, yu2023mimictouch, shukla2025LearningForceConditionedVisuomotor, mozaffari2025learning}. Recent works like Force-Aware Reactive Policy (FoAR) \cite{he2025foar} propose reactive policies for contact-rich tasks but often lack the hierarchical planning capability necessary for multi-stage disassembly. Recent advancements have further specialized these concepts for material removal and forceful disassembly \cite{haddadin2024unified}. In the domain of robotic cutting and chiseling, TopoCut \cite{topo_cut_25} introduces a framework for multi-step cutting that integrates differentiable damage tracking. In contrast, our approach aims to learn these high-frequency, force-conditioned behaviors directly from demonstrations within a hierarchical framework.

In the broader context of manufacturing, literature has increasingly focused on integrating learning-based control with digital twins for disassembly. Evaluating the fidelity of digital twins for robotic disassembly processes \cite{han2025digital, qin2026robot} has become critical for ensuring that learned policies transfer to physical hardware. Recent systems for automated screw removal \cite{diaz2025robotic} and battery disassembly \cite{ricard2024design} leverage these high-fidelity models to handle the variability of End-of-Life products. Furthermore, hierarchical learning frameworks \cite{simonivc2024hierarchical} have successfully combined high-level task planning with low-level contact primitives, a structure that is similar to our DPA-FTG approach, but often lacks the generative flexibility of diffusion models for the low-level policy \cite{unger2024prosip}. Other works have explored human-robot collaborative disassembly \cite{hu2024ontology, zafar2024exploring} and defect detection \cite{el2025real}, establishing a strong precedent for using data-driven methods in unstructured manufacturing environments \cite{lu2025vla}.

\section{Problem Formulation}
\label{sec:Problem Formulation}

\noindent We consider the domain of contact-rich bimanual manipulation, utilizing a dual-arm system comprising of two tools e.g, chisel and gripper. We formulate the task as a Learning from Demonstration (LfD) problem within a Partially Observable Markov Decision Process (POMDP) framework.

\subsection{State, Action, and Observation Spaces}

Let the environment be defined by the tuple $\mathcal{M} = (\mathcal{S}, \mathcal{A}, \mathcal{O}, \mathcal{T})$.

\noindent \textbf{State Space ($\mathcal{S}$):} The full system state $s_t \in \mathcal{S}$ encompasses the kinematic states of both manipulators (joint angles $\mathbf{q}$, velocities $\dot{\mathbf{q}}$), the workpiece pose, and the hidden physical parameters of the environment. In material separation tasks, these hidden parameters include the spatially varying bond strength $\beta(\mathbf{x})$, local friction coefficients $\mu$, and the instantaneous contact mode (e.g., sticking vs. slipping). Since these parameters cannot be directly measured, the policy must infer them from the interaction history.

\noindent \textbf{Action Space ($\mathcal{A}$):} Let the action at time $t$ be denoted by $\mathbf{a}_t \in \mathcal{A}$. We define the action space as the domain of joint velocity commands:

\begin{equation}
\mathbf{a}_t = [ \dot{\mathbf{q}}^{chisel}_t, \dot{\mathbf{q}}^{gripper}_t ]^\top
\end{equation}

Velocity control is selected to facilitate compliant interaction. This allows the policy to naturally learn impedance-like behaviors (e.g. such as halting motion ($\mathbf{a}_t \approx 0$) or retracting ($\mathbf{a}_t < 0$)) upon encountering unmodeled rigid constraints, without requiring explicit torque control.

\noindent \textbf{Observation Space ($\mathcal{O}$):} The robot receives a multi-modal observation $\mathbf{o}_t$ at each timestep $t$. This observation comprises:
\begin{equation}
    \mathbf{o}_t = \{ \mathbf{I}_t, \mathbf{q}_t,  \mathbf{\dot q_t}, \mathbf{F}_t \}
\end{equation}
where $\mathbf{I}$ represents RGB images, $\mathbf{q}$ and $\mathbf{\dot q}$ denote proprioception, and $\mathbf{F} \in \mathbb{R}^6$ represents contact wrenches.

\noindent \textbf{Transition Dynamics ($\mathcal{T}$):} Let $\mathcal{T}(s_{t+1} | s_t, \mathbf{a}_t)$ denote the unknown state transition probability function. This governs the evolution of the physical system, e.g., material deformation and tool-surface friction. Since it is difficult to model analytically in contact-rich scenarios, we adopt a model-free learning approach.

\subsection{Learning Objective}

We assume access to a dataset of expert demonstrations $\mathcal{D} = \{ (\mathbf{o}_t, \mathbf{a}_t) \}_{t=0}^T$. The goal is to learn a policy $\pi_\theta$ that matches the expert's action distribution by maximizing the log-likelihood:

\begin{equation}
\max_\theta \mathbb{E}_{(\mathbf{o}, \mathbf{a}) \sim \mathcal{D}} [ \log \pi_\theta(\mathbf{a}_t | \mathbf{o}_{0:t}) ]
\end{equation}

A fundamental challenge in this domain is the mismatch between the timescale of generative inference and the timescale of physical interaction. Standard diffusion policies typically rely on Action Chunking, predicting a sequence of actions $\mathbf{a}_{t:t+T_a}$ and executing them in an open-loop. Let $f_{env}$ be the characteristic frequency of contact dynamics (e.g., fracture propagation or rigid impact). In chiseling, $f_{env} \approx 60 \text{ Hz}$, while diffusion inference typically operates at $f_{inf}\approx 5 \text{ Hz}$.

We identify this as the Open-Loop Commitment failure mode. Executing a pre-planned trajectory open-loop for the duration of the inference window ($1/f_{inf} \approx 200$ ms) renders the system blind to high-frequency contact events. If the tool encounters a rigid knot during this window, the committed velocity profile results in unbounded force growth, leading to hardware damage or tool failure. Therefore, the desired policy should satisfy the following constraints:

\begin{enumerate}
\item \textbf{Asymmetric Coordination:} Coordinate the Chisel arm (applying separation force) and Gripper arm (stabilizing the object) simultaneously;
\item \textbf{High-Frequency Reactivity:} React in real-time to force feedback; if a sudden resistance spike is detected, the policy must modulate the action (e.g., retract or oscillate) within milliseconds;
\item \textbf{Failure Recovery:} Recover from local failures autonomously, such as re-grasping if the sheet slips;
\item \textbf{Generalization:} Adapt to variations in object pose or orientation without re-calibration; 
\item \textbf{Task Multimodality:} Select appropriate task modes, (e.g. switching between chiseling, prying, or peeling) based on the multi-modal context.
\end{enumerate}

\section{Approach}
\label{sec:approach}

\noindent As discussed in Sec. \ref{sec:Problem Formulation}, forceful manipulation (e.g., chiseling) requires high-bandwidth feedback control (typically $f_{\text{ctrl}}\approx 60$ Hz) to remain stable under rapidly changing contact forces. In contrast, diffusion-based policies require multiple denoising iterations and are therefore best suited for lower-frequency planning ($f_{\text{plan}}\approx 5$ Hz). To address the frequency mismatch between high-frequency contact dynamics and low-frequency policy inference, we propose \textbf{Diffusion Policy Augmented by Fast Trajectory Generation (DPA-FTG)}, that (i) uses diffusion for slow task selection and (ii) uses a lightweight recurrent controller for fast force-reactive trajectory execution.

\subsection{Overview of Approach}
\label{subsec:approach_overview_approach}

\begin{figure*}[p]
    \centering
    \includegraphics[width=\linewidth]{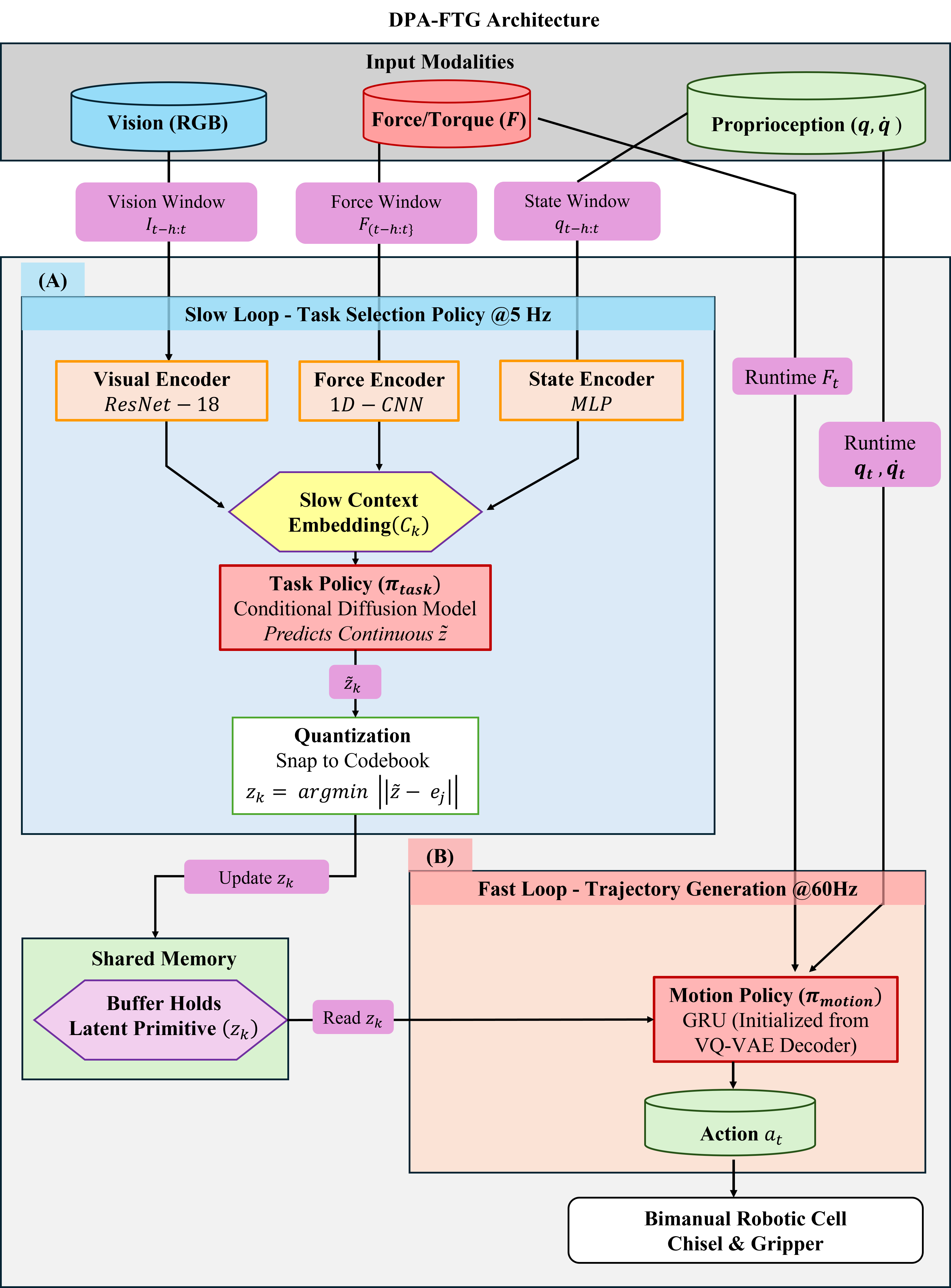}
     \caption{
        \textbf{System Architecture of DPA-FTG.}
        The framework decouples high-level task planning from low-level reactive control.
        (A) Slow Loop (5 Hz): The Task Selection Policy ($\pi_{task}$) takes a history of multi-modal observations (Vision, Force, Proprioception) to predict a continuous latent embedding $\tilde{z}$. This embedding is quantized to the nearest codebook vector $z_k$ and published to an atomic buffer.
        (B) Fast Loop (60 Hz): The Motion Generation Policy ($\pi_{motion}$), running on a separate thread, reads the latest skill $z_k$. It combines this semantic intent with instantaneous force and proprioceptive feedback to generate high-frequency velocity commands ($\mathbf{a}_t$). The fast loop operates independently of the planner's latency, ensuring continuous 60 Hz reactivity.
     }
    \label{fig:DPA-FTG}   
\end{figure*}

DPA-FTG shifts the concept of chunking from the action space to a latent task space. We posit that while the execution of a task must be high-frequency and reactive, the intent of that task remains stable over longer horizons. Instead of diffusing a rigid sequence of poses, our High-Level Policy ($\pi_{task}$) predicts a Latent Primitive ($z_k$). This primitive encodes the semantic intent of the trajectory at timestep $k$, (e.g., "Oscillate forward 2cm" versus "Pry upward") without committing to the specific motor commands required to achieve it.

This stable latent task intent is passed to the low-level Force-Reactive Motion Decoder ($\pi_{motion}$), implemented as a Gated Recurrent Unit (GRU). Unlike standard decoders that reconstruct a fixed, open-loop action sequence, our policy unzips the latent intent ($z_k$) into actions ($a_t$) autoregressively. By generating actions step-by-step at 60Hz, the system can modulate the trajectory inside the chunk. If a force spike is detected at step $t$, the decoder can instantaneously deviate from the nominal path (e.g., retracting velocity) at step $t+1$ without waiting for the next high-level re-plan.

Formally, DPA-FTG decouples the policy into two temporal abstractions: (1) a Task Selection Policy ($\pi_{task}$) that operates at a coarse frequency to perform task selection, and (2) a Motion Generation Policy ($\pi_{motion}$) that operates at a high frequency to execute said tasks with closed-loop force feedback. The interface between these two policies is a learned latent primitive space $\mathcal{Z}$, which bridges high-level task selection and low-level execution. We aim to learn a compact latent space $\mathcal{Z}$ where each vector $z \in \mathcal{Z}$ encapsulates a specific motion skill (e.g., oscillatory chiseling vs. steady peeling). The overall system architecture is illustrated in Fig. \ref{fig:DPA-FTG}.

\noindent \textbf{Observation Space ($\mathcal{O}$):} To address the computational asymmetry between semantic planning and contact stability, we partition the observation space into two distinct temporal contexts. Let $t$ denote the high-frequency control step ($60$ Hz) and $k$ denote the low-frequency planning step ($5$ Hz).

\begin{itemize}

\item \textbf{Slow Task Context ($\mathbf{o}^{slow}_k$):} A subset of $\mathcal{O}$, used to infer the latent state to select task primitive. This context is sampled at step $k$ (corresponding to time $t$) and comprises a history window of length $T_h$:

\begin{equation}
\label{eq:overview_slow_obs_context}
\mathbf{o}^{slow}_k = \{ \mathbf{I}_{t_k-T_h:t_k}, \mathbf{q}_{t_k-T_h:t_k}, \mathbf{F}_{t_k-T_h:t_k} \}
\end{equation}

where $\mathbf{I}$ represents RGB image sequences, $\mathbf{q}$ represents the history of joint positions, and $\mathbf{F} \in \mathbb{R}^{T_h \times 6}$ represents the contact force history.

\item \textbf{Fast Reactive Context ($\mathbf{o}^{fast}_t$):} A subset of $\mathcal{O}$, used for real-time stabilization. This context captures only the instantaneous state at time $t$ required for closed-loop control:
\begin{equation}
\label{eq:overview_fast_obs_context}
    \mathbf{o}^{fast}_t = \{ \mathbf{q}_t, \mathbf{\dot q_t}, \mathbf{F}_t \}
\end{equation}

\end{itemize}

We define the control law as:
\begin{equation}
\label{eq:overview_control_law}
\mathbf{a}_t = \pi_{motion}(\mathbf{z}_k, \mathbf{o}^{fast}_t)
\end{equation}

where $\mathbf{z}_k$ is a discrete latent variable generated by a high-level planner:

\begin{equation}
\label{eq:overview_z_k}
\mathbf{z}_k \sim \pi_{task}(\cdot | \mathbf{o}^{slow}_k)
\end{equation}

Here, $k = \lfloor t \cdot (f_{plan}/f_{ctrl}) \rfloor$ indexes the coarse planning steps ($f_{plan} \approx 5$ Hz), while $t$ indexes the fine control steps ($f_{ctrl} = 60$ Hz). $\pi_{task}$ acts as the task selector, while $\pi_{motion}$ acts as a closed-loop decoder that unrolls the task intent $\mathbf{z}_k$ into high-frequency velocity commands.

\noindent The realization of this hierarchical formulation imposes three specific architectural constraints, which we discuss in the following sections \ref{subsec:approach_learning_z}-\ref{subsec:appraoch_async_dep}. 

First, the latent interface $\mathcal{Z}$ cannot be arbitrarily continuous. Since contact strategies (e.g., "pry" vs. "cut") are often mutually exclusive, a continuous latent space would encourage invalid mode-averaging. To resolve this, Sec. \ref{subsec:approach_learning_z} details the construction of a discrete vocabulary of motion primitives using a Vector Quantized Variational Autoencoder (VQ-VAE) \cite{NIPS2017_7a98af17}.

Second, determining the optimal primitive $z_k$ is a generative problem (Eq. \ref{eq:overview_z_k}), not a discriminative one. In a given state of the system, the distribution of valid strategies is highly multi-modal. To capture this diversity without collapsing to the mean, Sec. \ref{subsec:approach_high_diffusion} discusses $\pi_{task}$ as a Conditional Diffusion Model. This allows the planner to sample valid intents from the learned codebook based on the slow context $\mathbf{o}^{slow}_k$.

\noindent Third, the control law (Eq. \ref{eq:overview_control_law}) requires the policy to generate complex kinematic skills (defined by $z_k$) while simultaneously reacting to high-frequency feedback ($\mathbf{o}^{fast}_t$). Learning this dual capability from scratch is difficult. Sec. \ref{subsec:approach_low_gru} describes our transfer learning strategy, where $\pi_{motion}$ inherits the kinematic structure of the VQ-VAE decoder and is fine-tuned to become a closed-loop controller, enabling it to modulate execution speed and direction in real-time based on contact forces.

\noindent Finally, Section \ref{subsec:appraoch_async_dep} presents the asynchronous deployment architecture that physically bridges the $f_{plan}$ and $f_{ctrl}$ timescales defined in section \ref{sec:Problem Formulation}.

\subsection{Learning the Latent Primitive Space}
\label{subsec:approach_learning_z}

\begin{figure*}[t]
    \centering
    \includegraphics[width=\linewidth]{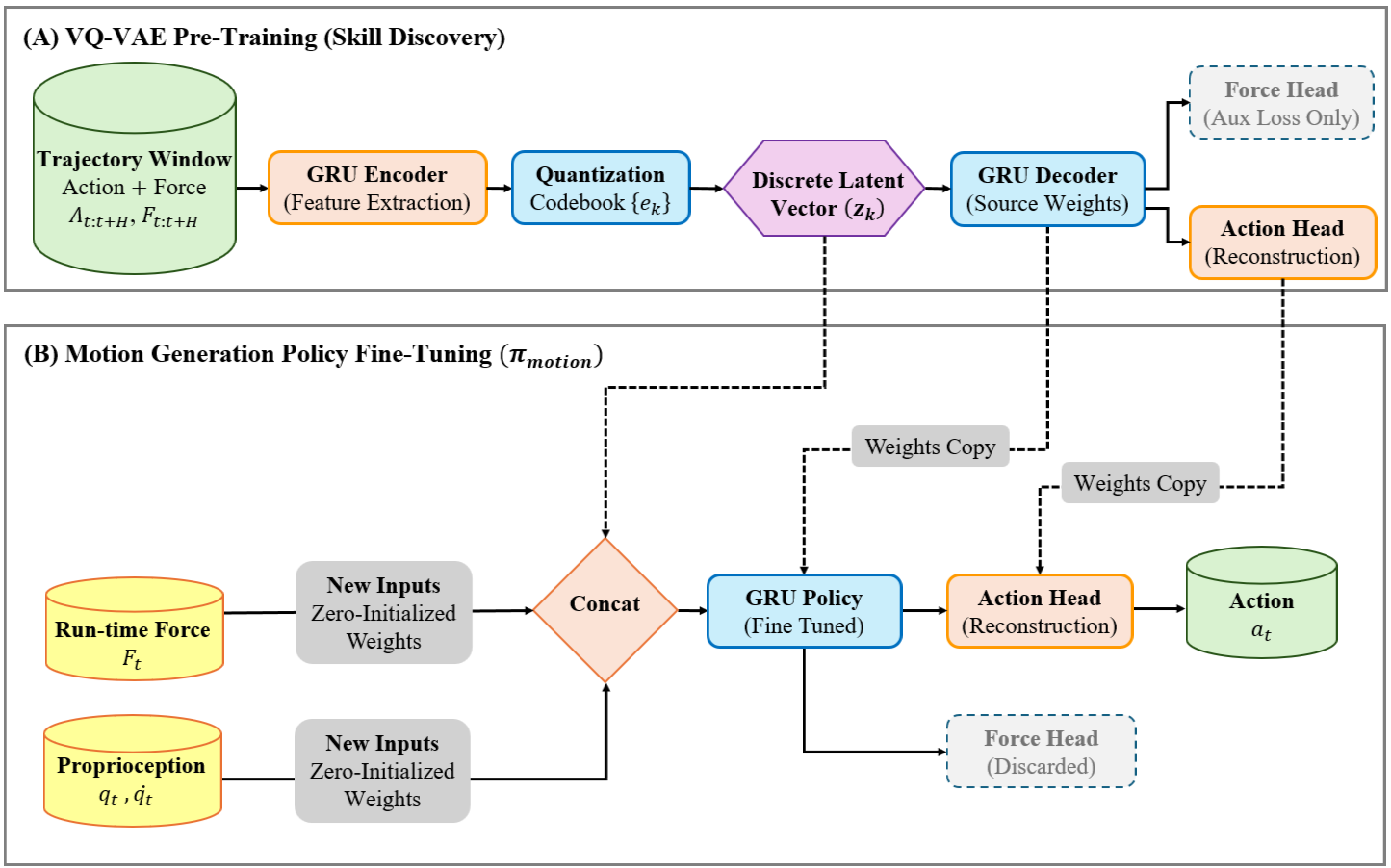} 
    \caption{\textbf{Architecture for Latent Learning and Policy Transfer.} (A) Skill Discovery (VQ-VAE): During pre-training, the encoder compresses expert trajectories (Action + Force) into a discrete latent code $z_q$. The decoder is trained to reconstruct the action profile $\hat{\mathbf{a}}$, with an auxiliary head reconstructing the force profile $\hat{\mathbf{f}}$ to ensure the latent captures dynamic context. (B) Policy Fine-Tuning: The low-level controller ($\pi_{motion}$) is initialized by transferring weights from the pre-trained VQ-VAE decoder (blue arrows). To enable reactivity, a new input layer is grafted onto the network to accept real-time sensor data ($\mathbf{F}_t, \mathbf{q}_t$). These new weights are zero-initialized to preserve the pre-trained kinematic behavior at the start of fine-tuning. The auxiliary force head is discarded, and the policy is trained via denoising to output actions conditioned on both the latent skill and runtime force observations.}
    \label{fig:vae_arch}
\end{figure*}

To learn the task primitive $\mathcal{Z}$, we require a method that compresses high-dimensional expert trajectories into a compact, discrete vocabulary of skills. We achieve this by training a VQ-VAE \cite{NIPS2017_7a98af17}. A standard VAE maps trajectories to a continuous Gaussian distribution. In a multimodal task like disassembly, where the robot must sharply choose between distinct strategies (e.g., "Pry Left" vs. "Pry Right"), a Gaussian VAE tends to collapse these task primitives.  This results in averaged behaviors that are physically invalid. By using a VQ-VAE we learn a discrete vocabulary of skills with a discrete codebook $\mathcal{C} = \{e_k\}_{k=1}^K$, we force the system to commit to specific, distinct task primitives, preventing mode averaging. This adopts the Latent Diffusion Model (LDM) paradigm \cite{rombach2022high}, shifting the generative process from the high-dimensional raw action space to a compressed, semantically rich latent manifold.

\textbf{Trajectory Representation:} We define a trajectory window of horizon $H$ as

$$\tau_t = \{ (\mathbf{a}_i, \mathbf{f}_i) \}_{i=t}^{t+H},$$

where $\mathbf{a}_i$ represents the joint velocity commands and $\mathbf{f}_i$ represents the contact forces. Including force profiles in the VQ-VAE input is a critical design choice for representation learning. Kinematically, "moving in free space" and "pushing against a bond" may appear identical in velocity space. By conditioning the encoder on force, we ensure that the learned primitives capture the dynamic context of the task, distinguishing between free-motion and contact-rich interaction.

\textbf{Architecture and Quantization:} The VQ-VAE consists of an encoder $\mathcal{E}_\phi$, a discrete codebook $\mathcal{C} = \{ \mathbf{e}_k \}_{k=1}^K \subset \mathbb{R}^{D_z}$, and a decoder $\mathcal{D}_\psi$. The encoder maps the input window $\tau_t$ to a continuous latent vector $\mathbf{z}_e(\tau_t)$. This vector is then discretized by "snapping" it to the nearest vector in the codebook:

\begin{equation}
\label{eq:quantization}
\mathbf{z}_q = \mathbf{e}_k \quad \text{where} \quad k = \arg \min_j || \mathbf{z}_e(\tau_t) - \mathbf{e}_j ||_2
\end{equation}

This quantization step creates a discrete bottleneck, forcing the model to commit to one of $K$ distinct motion primitives rather than interpolating between them.

\textbf{Training Objective:} The decoder $\mathcal{D}_\psi$ takes the quantized primitive $\mathbf{z}_q$ and attempts to reconstruct the original trajectory. To ensure the latent space encodes both kinematic intent and dynamic context, we employ a multi-objective loss function:

\begin{equation}
\label{eq:vq_loss}
\mathcal{L}_{VQ} = \underbrace{\mathcal{L}_{rec}(\mathbf{a}, \hat{\mathbf{a}})}_{\text{Action Recon.}} + \underbrace{\lambda \mathcal{L}_{rec}(\mathbf{f}, \hat{\mathbf{f}})}_{\text{Aux. Force Recon.}} + \underbrace{\| \text{sg}[\mathbf{z}_e] - \mathbf{e}_k \|_2^2}_{\text{Codebook Loss}} + \underbrace{\beta \| \mathbf{z}_e - \text{sg}[\mathbf{e}_k] \|_2^2}_{\text{Commitment Loss}}
\end{equation}

where $\text{sg}[\cdot]$ denotes the stop-gradient operator. The first term ensures kinematic fidelity. The second term is an auxiliary loss that forces the latent $\mathbf{z}_q$ to retain information about contact dynamics. Although we do not use the predicted force $\hat{\mathbf{f}}$ for control during deployment, this auxiliary objective forces the latent code $z$ to encode information about the expected contact dynamics, ensuring the decoder knows the difference between stiff and compliant interactions. This objective prevents the encoder from ignoring the force inputs. The final two terms update the codebook and constrain the encoder output to the codebook manifold, respectively.

This pre-training stage effectively compresses the continuous, high-frequency expert data into a sequence of discrete tokens $\mathbf{z}_q$ (see Fig. \ref{fig:vae_arch}), providing a stable, lower-dimensional manifold for the high-level planner.

\subsection{High-Level Task Selection Policy}
\label{subsec:approach_high_diffusion}

The role of the high-level policy $\pi_{task}$ is to select the appropriate primitive $z_k$ based on the visual and physical state. Unlike standard diffusion policies that output sequences of end-effector poses, $\pi_{task}$ predicts the latent primitive $z_k$ required to progress the task over the next horizon $H$. We model this policy as a conditional diffusion model DDPM \cite{chi2023diffusion}. Unlike a classification network that outputs a single probability distribution, diffusion models are generative. This allows the planner to represent the full distribution of valid strategies, which is essential in ambiguous states where multiple valid recovery behaviors might exist. A simple classifier often struggles from mode collapse, whereas diffusion preserves the diversity of expert strategies.

\textbf{Training supervision.} During training, we treat the codebook vector $e_{n_k}$ (associated with the expert's demonstration at slow step $k$ via the VQ-VAE encoder) as the ground-truth \emph{clean} target in the skill embedding space. The diffusion model is trained with the standard DDPM noise-prediction objective so that, conditioned on the slow context, it denoises noisy versions of $e_{n_k}$ back towards $e_{n_k}$.

A conditional diffusion process models the distribution $p(\tilde{z}_k \mid \mathbf{o}^{slow}_k)$ by reversing a forward noising process in the continuous embedding space. Starting from Gaussian noise $z^{(K)} \sim \mathcal{N}(0, I)$, the policy iteratively denoises the latent vector over $K$ steps. The update rule at reverse step $i$ is:

\begin{equation}
\label{eq:subsec_high_conti_diffusion}
    z^{(i-1)} = \alpha_i \bigl(z^{(i)} - \gamma_i \,\epsilon_\theta(z^{(i)}, i, \mathbf{o}^{slow}_k) \bigr) + \sigma_i \omega,
\end{equation}

where $\epsilon_\theta$ is the learned noise-prediction network, $i$ indexes the reverse diffusion steps, and $\mathbf{o}^{slow}_k$ represents the current multi-modal context (vision, force history, and proprioception).

After $K$ reverse steps, we obtain the final denoised sample
$$
\tilde{z}_k = z^{(0)} \in \mathbb{R}^{D_z},
$$
which is a continuous embedding in the same space as the codebook vectors.

The model predicts a continuous embedding vector $\tilde{z}$. Crucially, the low-level controller is only trained to interpret valid skills drawn from the VQ-VAE codebook $\mathcal{C} = \{e_j\}$. To map the continuous prediction $\tilde{z}_k$ back onto this discrete skill manifold, we perform a nearest-neighbor quantization:

\begin{equation}
\label{eq:subsec_high_task_nearest_search}
z_k = \arg\min_{e \in \mathcal{C}} \| \tilde{z}_k - e \|_2
\end{equation}

Here $z_k$ is the codebook vector (skill embedding) that will be written into the shared latent buffer and used by the motion policy \ref{subsec:appraoch_async_dep}.
This quantization acts as a filter, ensuring that the low-level controller only ever receives a valid, recognizable skill token $e \in \mathcal{C}$, effectively removing any residual variance from the continuous diffusion inference.

The observation $\mathbf{o}^{slow}_k$ consists of RGB images from global and wrist-mounted cameras, processed by a frozen ResNet-18 backbone to extract spatial features, and the force history $\mathbf{F}_{t-T_h:t}$ processed by a 1D-CNN encoder. This temporal aggregation allows the planner to identify long-horizon semantic states (e.g., distinguishing a "stuck" tool from a "hovering" one based on force trends) while ignoring the transient spikes handled by the low-level reflex controller. These features are concatenated with the robot's current joint configuration.

These features are concatenated with the robot's current joint configuration and used to condition the U-Net denoising backbone using Feature-wise Linear Modulation (FiLM) \cite{chi2023diffusion}. By diffusing in the continuous embedding space of $\mathcal{Z}$, $\pi_{task}$ effectively selects the optimal task primitive (e.g., Oscillate-Insertion, Lever-Pry, Peel-Back).

\subsection{Low-Level Motion Generation Policy}
\label{subsec:approach_low_gru}

The Low-Level Policy, $\pi_{motion}(a_t | z_k, \mathbf{O}_t^{fast})$, functions as a high-frequency trajectory generator. It executes the semantic skill $z_k$ at 60 Hz while actively modulating motion based on real-time contact dynamics.

We implement $\pi_{motion}$ as a GRU \cite{xue_reactive_2025}. To avoid training from scratch, we initialize the policy weights directly from the pre-trained VQ-VAE Decoder. This ensures the policy begins with a perfect understanding of kinematic skills. However, two structural modifications are required to convert the open-loop decoder into a closed-loop policy:

\begin{enumerate}
    \item \textbf{Head Selection:} The VQ-VAE decoder outputs both actions and forces (for representation learning). Since $\pi_{motion}$ is strictly a controller, we retain the Action Head but discard the Force Reconstruction Head.
    \item \textbf{Input Injection via Zero-Initialization:} The original decoder accepts only the latent $z_k$. We expand the GRU's input layer to accept instantaneous force $\mathbf{F}_t$ and proprioception $\mathbf{q}_t, \dot{\mathbf{q}}_t$. Crucially, the weights connecting these new sensory inputs to the hidden state are initialized to zero.
\end{enumerate}

We copy the original decoder's GRU input weights for the $z_k$ channels, leave the hidden-to-hidden weights unchanged, and only the newly added input channels $(\mathbf{F}_t,\mathbf{q}_t,\dot{\mathbf{q}}_t)$ start at zero, so the initial behavior exactly matches the pre-trained open-loop decoder.

This zero-initialization strategy ensures that at the very first step of fine-tuning, the contribution of the new sensors is null, and the policy behaves exactly like the stable, pre-trained decoder. As training progresses, the network gradually updates these weights to incorporate feedback.

\noindent \textbf{Closed-Loop Control Law:} 
The policy unrolls the latent intent $z_k$ (held constant for the planning window) into motor commands.
The control law is defined in two stages: the GRU updates its hidden state from the latent and fast observations, and an action head maps the hidden state to a command:
\begin{equation}
    h_t = \text{GRU}_{\pi}([z_k, \mathbf{F}_t, \mathbf{q}_t, \dot{\mathbf{q}}_t], h_{t-1}), \qquad
    a_t = \text{MLP}_{\pi}(h_t).
\end{equation}

We explicitly include proprioception ($\mathbf{q}_t, \dot{\mathbf{q}}_t$) to ensure the controller allows for state-dependent modulation. The recurrent hidden state $h_{t-1}$ maintains the phase of periodic motions (e.g., oscillation).

A key design strength of DPA-FTG is the specific partition of inputs. The high-level planner consumes history of features (Vision/Force history) to select complex contact states over time. In contrast, $\pi_{motion}$ consumes only instantaneous feedback ($\mathbf{F}_t, \mathbf{q}_t$). This removes the computational overhead of processing history buffer of high dimensional vision signal in the 60 Hz loop, minimizing input latency for critical contact reactions.

\noindent \textbf{Implicit Admittance via Denoising:} 
A recurrent network trained purely on successful, smooth expert demonstrations tends to ignore force inputs, collapsing into an open-loop driver. To counter this, we treat the fine-tuning of $\pi_{motion}$ as a Denoising Autoencoder for Dynamics. 

During training, we inject Gaussian noise and small perturbations into the fast inputs, e.g.,
$$
\mathbf{F}_{\text{noisy}} = \mathbf{F}_t + \epsilon_F,\quad
\mathbf{q}_{\text{noisy}} = \mathbf{q}_t + \epsilon_q,\quad
\dot{\mathbf{q}}_{\text{noisy}} = \dot{\mathbf{q}}_t + \epsilon_{\dot{q}},
$$
and feed $[z_k, \mathbf{F}_{\text{noisy}}, \mathbf{q}_{\text{noisy}}, \dot{\mathbf{q}}_{\text{noisy}}]$ into the GRU, while still supervising the clean expert actions.

We use a simple reconstruction loss over actions,
\begin{equation}
    \mathcal{L}_{\text{motion}} = \sum_t \big\| a_t - a_t^{\text{expert}} \big\|_2^2.
\end{equation}

The result is a learned implicit admittance-like behavior: the latent $z_k$ sets the nominal motion profile, and deviations in the sensed forces cause the policy to adjust its velocity output to maintain contact stability (see Fig. \ref{fig:vae_arch}).

\subsection{Asynchronous Inference and Deployment}
\label{subsec:appraoch_async_dep}

\begin{figure*}[t]
    \centering
    \includegraphics[width=\linewidth]{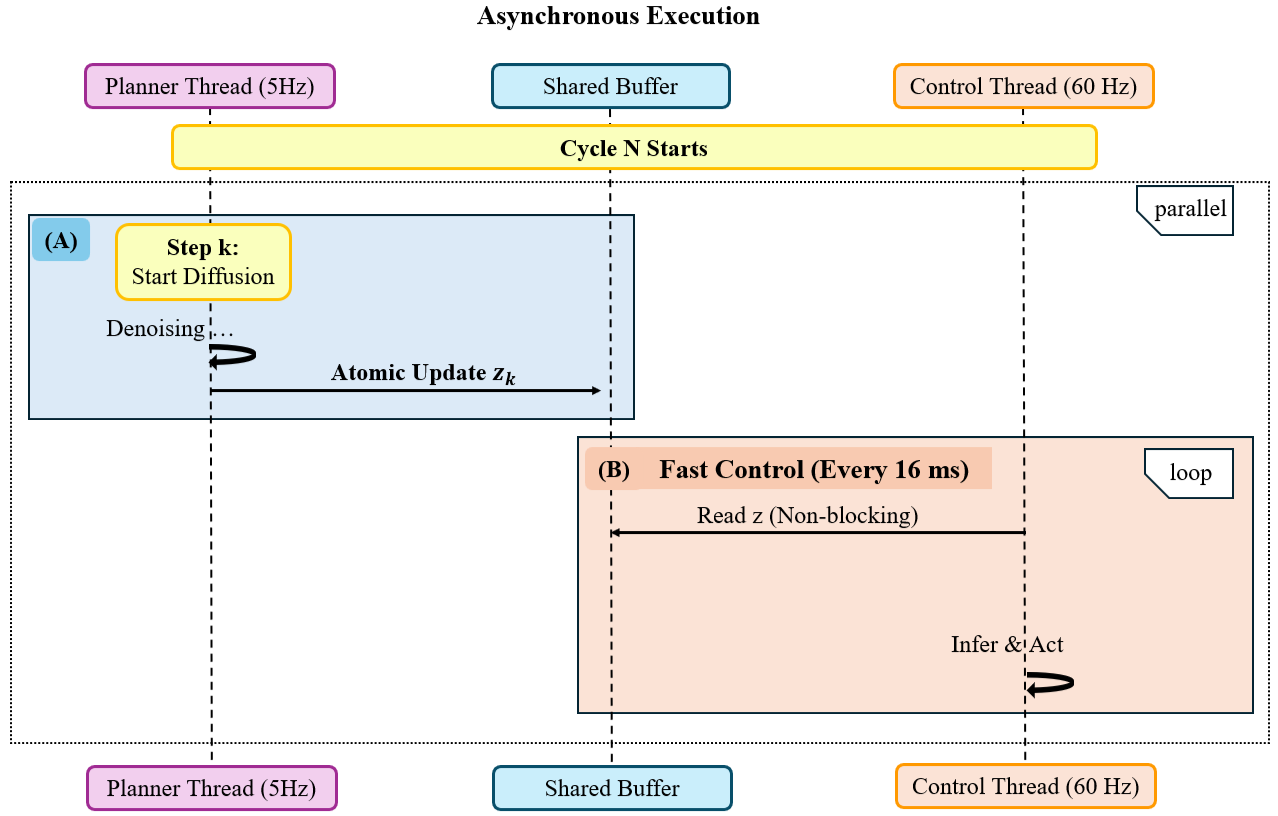} 
    \caption{\textbf{Asynchronous Inference Timing.} The system enables high-frequency control despite the computational latency of the diffusion planner. (A) Planner Thread: The diffusion process takes $\approx 200$ ms to denoise and predict the next target skill $z_k$. (B) Control Thread: Executing at 60 Hz (16 ms/tick), the controller reads the current target from the shared buffer in a non-blocking manner. This allows the robot to react to contact dynamics continuously without pausing for the planner.}
    \label{fig:timing_async}
\end{figure*}

DPA-FTG performs asynchronous execution of the two policies (see Fig. \ref{fig:timing_async}), which resolves the stop-and-go latency inherent in standard diffusion policies. The system architecture decouples the decision-making horizon from the control horizon via two parallel threads:

\begin{enumerate}
    \item \textbf{Planner Thread (5 Hz):} This thread runs the iterative denoising process for $\pi_{task}$. Computing the next target latent $z_k$ takes approximately $200$ ms. Upon completion, it updates a thread-safe shared atomic buffer.
    \item \textbf{Control Thread (60 Hz):} This thread executes the high-frequency control loop. At each timestep $t$, it reads the most recent latent $z_{\text{curr}}$ from the atomic buffer and the real-time sensors. It then executes a single forward pass of $\pi_{motion}$ to generate the immediate motor command.
\end{enumerate}

This Zero-Order Hold strategy ensures that the fast control loop is never blocked by the inference latency of the diffusion planner. The controller simply continues executing the last known valid intent until a new strategy is published.

\section{Results}
\noindent We designed a series of experiments to validate the performance of the proposed DPA-FTG architecture. Our evaluation focuses on three key questions: (1) Does the hierarchical architecture of task selection and trajectory generation improve performance over standard diffusion policy? (2) How does the system compare against strong baselines during execution in terms of safety? (3) Can the learned policy generalize to unseen geometries in a zero-shot manner?

\subsection{Experimental Setup}
\subsubsection{Robot Cell Design}
\begin{figure*}[t]
    \centering
    \includegraphics[width=\linewidth]{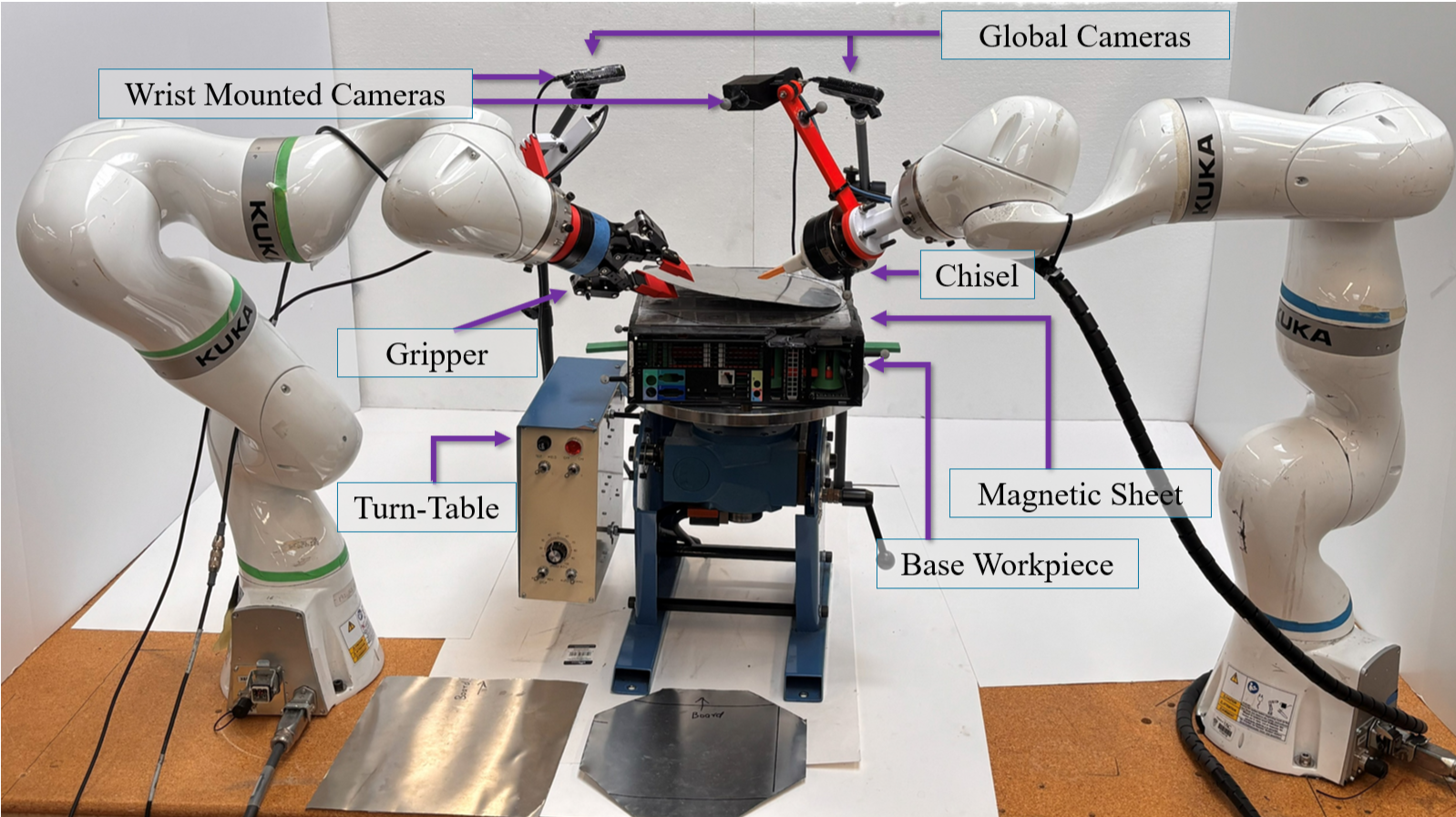}
     \caption{The bimanual robotic cell used for experimental validation. The left arm is equipped with a chisel and a force sensor to perform high-frequency oscillatory motions. The right arm is equipped with a gripper for peeling the separated sheet. The workpiece is mounted on a turntable to allow access to all sides.}
    \label{fig:Cell_setup}   
\end{figure*}

We validate our approach on a bimanual robotic cell consisting of two KUKA LBR iiwa 7 manipulators (Fig.~\ref{fig:Cell_setup}). One arm is equipped with a custom rigid chisel and the second arm utilizes a parallel gripper based on the Universal Manipulation Interface (UMI) design \cite{chi2024universal} to grasp and peel the sheet. To evaluate the system, we developed a surrogate workpiece that models the structure of a battery pack while ensuring experimental repeatability. To mitigate kinematic reach constraints of the robots, the workpiece is mounted on a motorized turntable. 

\begin{figure*}[t]
    \centering
    \includegraphics[width=\linewidth]{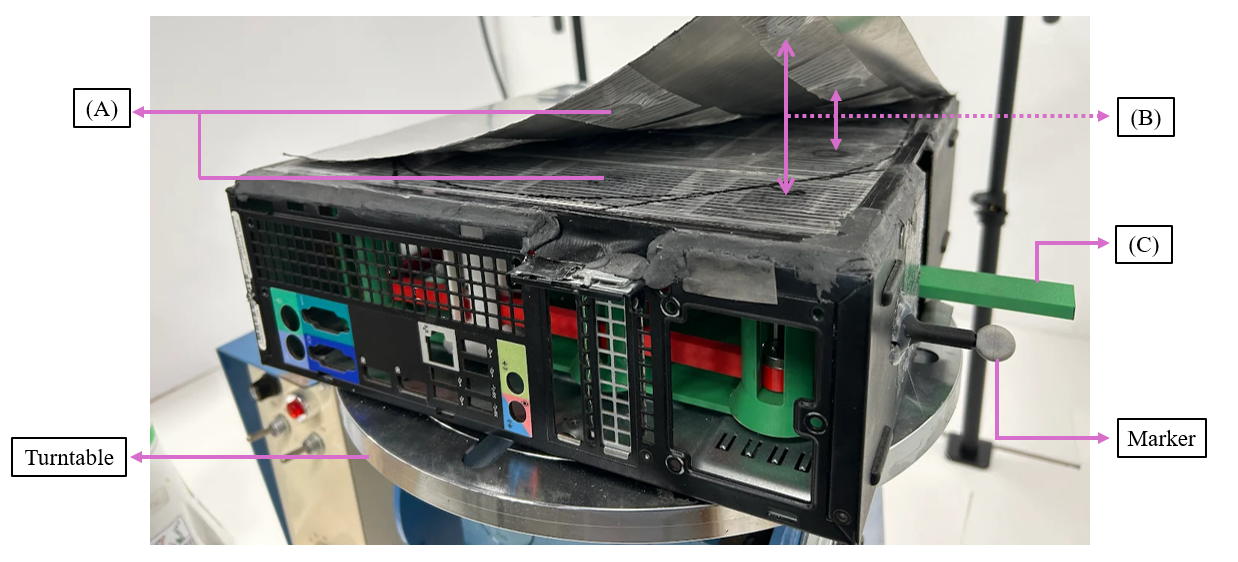}
     \caption{\textbf{Surrogate Workpiece Design and Magnetic Adhesion Mechanism.} The custom fixture models a battery module with a compliant aluminum cover. (A) Adhesion is provided by an array of 16 neodymium magnets. (B) Upon separation, magnets retract to simulate bond fracture. (C) A lever mechanism allows for rapid, repeatable re-engagement of magnets for subsequent trials.}
    \label{fig:workpiece_design}   
\end{figure*}

The workpiece consists of a compliant aluminum sheet adhered to a rigid base substrate. To simulate adhesive bonding while allowing for rapid, consistent resetting between trials, we utilize a magnetic fixture (see Fig.~\ref{fig:workpiece_design}). The sheet is secured by an array of 16 neodymium magnets evenly distributed along the bonding interface. The fixture design allows individual magnets to disengage and slide downwards once the local peeling force overcomes magnetic attraction, effectively simulating the progressive fracture of a chemical adhesive bond. A manual lever mechanism enables the simultaneous re-engagement of all magnets after loading a fresh sheet, providing consistent initial adhesion conditions across all experiments. This surrogate is designed to emulate the progressive release events and force transients, such as stick-slip and intermittent release, that are necessary to test high-frequency reactive control and task switching behavior. This experimental setup is not intended to exactly reproduce the full constitutive behavior of a specific chemical adhesive, thereby abstracting away complex material properties including strong rate dependence, temperature effects, and spatially heterogeneous cure states. Instead, the magnetic fixture provides a controlled and repeatable approximation that strictly preserves the key control-relevant challenge: rapid, discontinuous changes in the contact forces during the separation process.

\begin{figure*}[t]
    \centering
    \includegraphics[width=\linewidth]{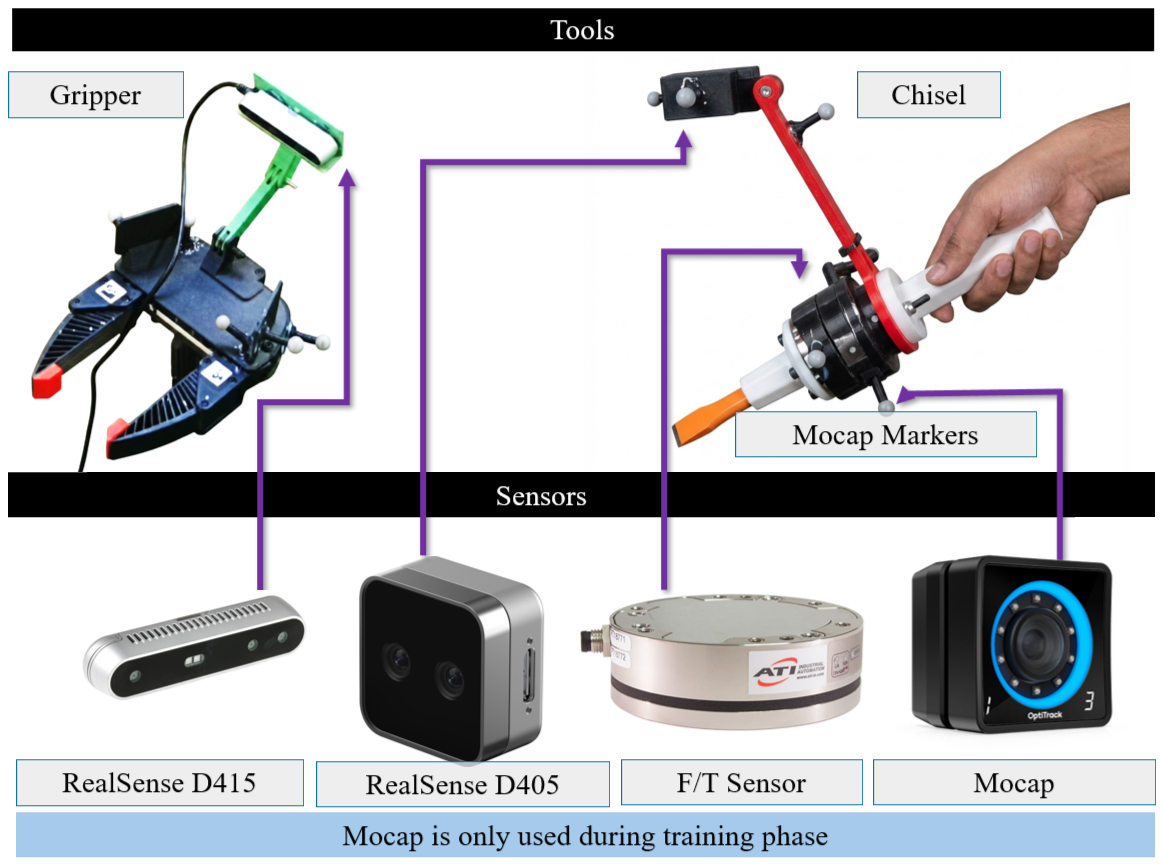}
     \caption{\textbf{Tools and Sensors.} (Top) Close-up views of the custom-designed gripper and chisel.  The gripper is based on the UMI-Gripper design \cite{chi2024universal}.
     (Bottom) The sensor suite comprises of two global RealSense D415 depth camera, an in-hand RealSense D405 depth camera, and an ATI Force sensor integrated with the chisel. Both tools are attached with retroreflective markers for MoCap. Note that the MoCap system is used exclusively during the training phase to acquire ground-truth trajectory data. During deployment, policy relies solely robot's onboard sensors for proprioception.}
    \label{fig:tools_n_sensors}   
\end{figure*}

The complete sensor and tool suite is detailed in Fig. \ref{fig:tools_n_sensors}. The chisel arm is equipped with an ATI Axia M20 Force sensor, capturing high-frequency contact dynamics at 1 kHz. Visual feedback is provided by a multi-view setup: (1) two global RealSense D415 cameras provide an overview of the workspace; (2) a wrist-mounted RealSense D405 camera on the gripper arm captures details of the peeling interface; and (3) a wrist-mounted RealSense D415 camera on the chisel arm ensures complete visual coverage of the tool-object interaction.

\begin{figure*}[t]
    \centering
    \includegraphics[width=\linewidth]{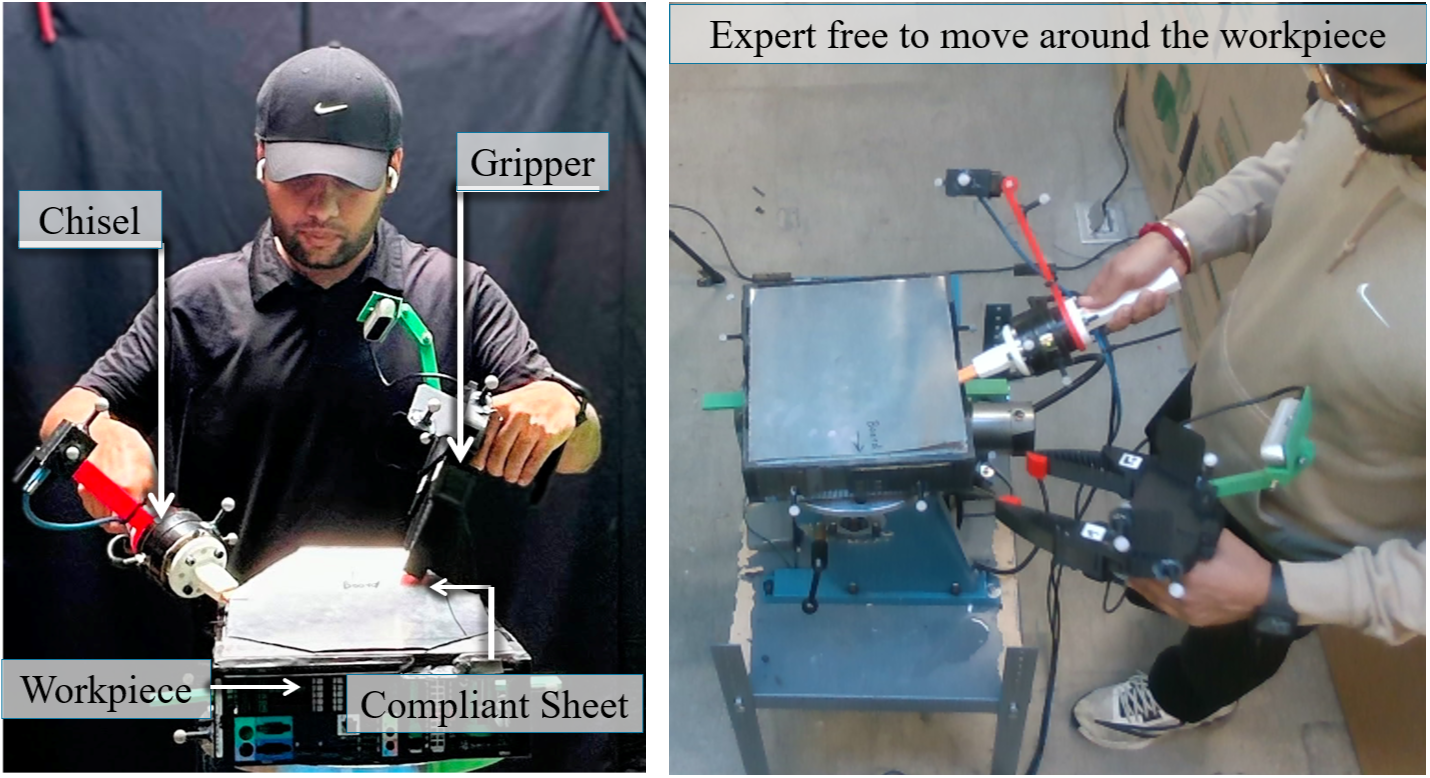}
     \caption{\textbf{Expert Demonstration Setup.} A human operator performs the bimanual compliant sheet separation task on a surrogate workpiece. This data collection method allows for the capture of natural manipulation dynamics, specifically the high-frequency oscillatory primitives required to break adhesive bonds, which are difficult to reproduce via teleoperation.}
    \label{fig:comparison_rigid_vs_deformable_swing}   
\end{figure*}

To acquire high-fidelity demonstrations, we prioritize direct human manipulation over teleoperation. While teleoperation is common in robotic imitation learning, it often introduces significant limitations for contact-rich, dynamic tasks, including communication latency, transparency loss in force feedback, and kinematic mismatch between the master and slave devices. These artifacts can filter out the high-frequency dynamics and subtle impedance modulations that are critical for breaking adhesive bonds.  By employing a Motion Capture (MoCap) system as shown in Fig. \ref{fig:comparison_rigid_vs_deformable_swing}, we allow the expert to manipulate the tools directly with their hands in a natural setting. This approach captures the uninhibited live dynamics of the human expert, ensuring that the training data reflects the true physical requirements of the task rather than the artifacts of a teleoperation interface. Crucially, this MoCap data is used exclusively during the training phase for proprioceptive supervision. The deployed policy relies only on the robot's onboard sensors for proprioception. 

For MoCap tracking, each tool is treated as a rigid body defined by a fixed cluster of retroreflective markers mounted on the tool handle away from the contact tip to reduce occlusion during interaction. In our setup, the chisel and the gripper each use five markers arranged in a non-coplanar configuration. This marker geometry is kept fixed across all demonstrations. We calibrate the rigid-body-to-tool transform once, and we then apply the same transform for all recorded demonstrations to obtain consistent 6-DoF tool poses.

\begin{figure*}[t]
    \centering
    \includegraphics[width=\linewidth]{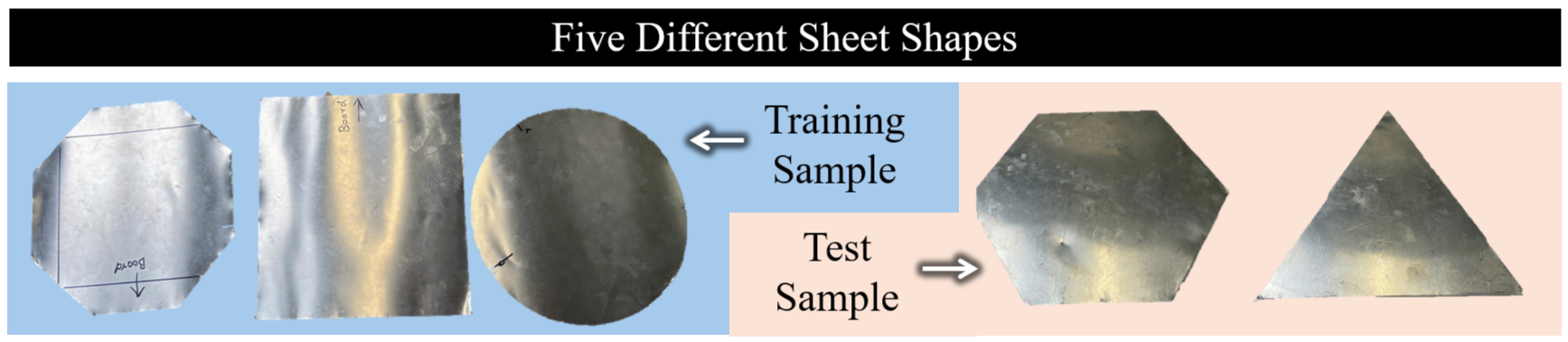}
     \caption{\textbf{Geometric Generalization Set.} The five distinct compliant sheet geometries used to evaluate the system. The Octagon, Square, and Circle serve as the training set. The Hexagon and Triangle are reserved as unseen test samples to evaluate the policy's zero-shot generalization to novel geometries.}
    \label{fig:5_sheets}   
\end{figure*}

To assess the policy's ability to generalize to novel object variations, we fabricated five distinct aluminum sheet geometries shown in Fig. \ref{fig:5_sheets}. The Octagon, Square, and Circle shapes comprise the training set. The Hexagon and Triangle shapes are used as an unseen test set to evaluate the policy's zero-shot geometric generalization capabilities.

\subsection{Data Collection}
The objective is the bimanual separation of the compliant sheet. The task comprises two coordinated phases: (i) Chiseling: the left arm performs high-frequency oscillatory chiseling along the bonded edges to break the adhesive seal, and (ii) Peeling: the right arm grasps and peels the separated sections of the sheet. This task was specifically chosen because high-frequency, force-controlled motion is essential for this instance, making it an ideal testbed for our method.

\begin{figure*}[t]
    \centering
    \includegraphics[width=\linewidth]{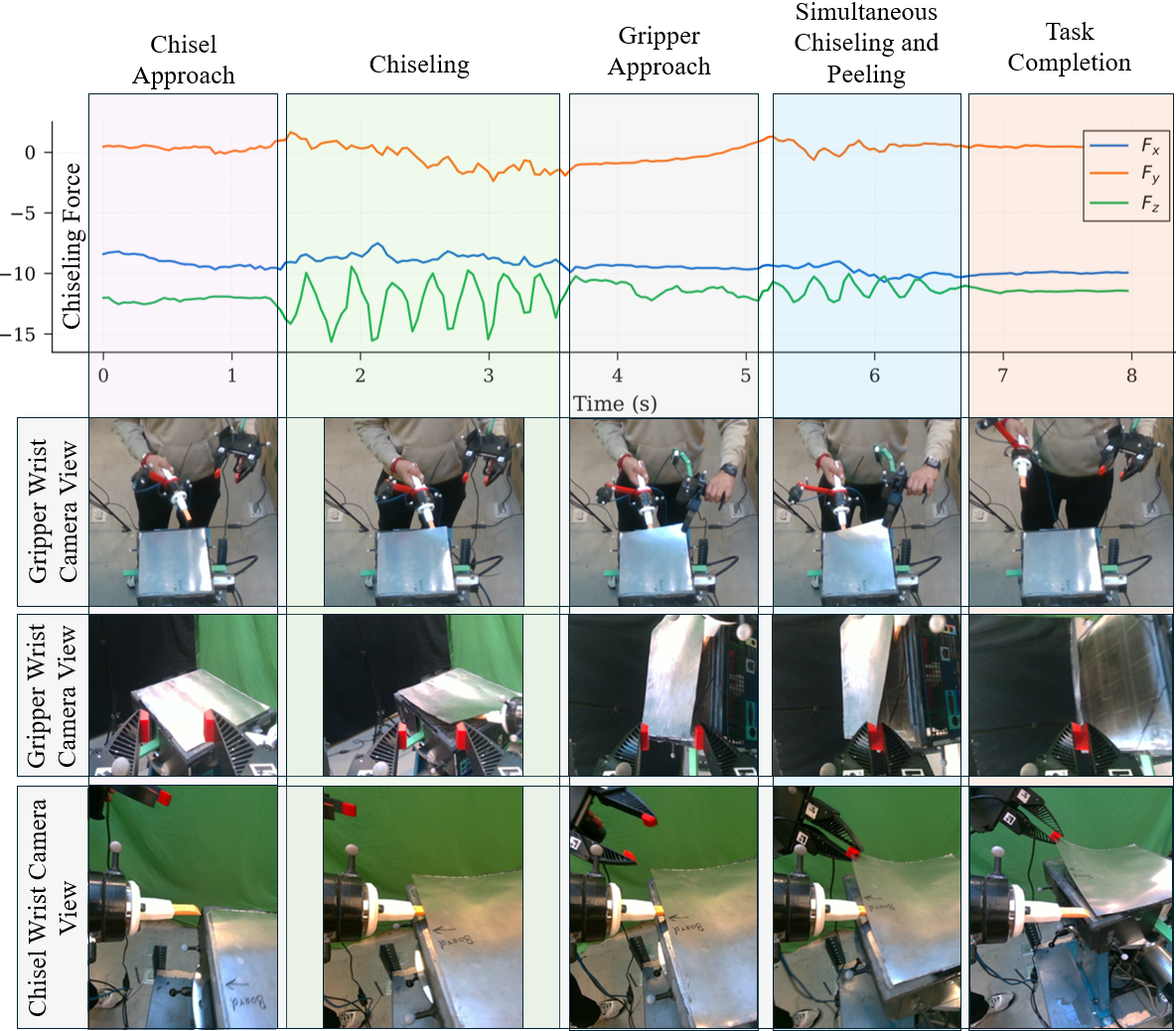} 
    \caption{A detailed breakdown of a human expert demonstration for the bimanual compliant sheet separation task, segmented into five distinct phases. The top panel displays the force profiles ($F_x, F_y, F_z$) measured at the chisel tool, highlighting the characteristic high-frequency oscillatory dynamics required during the "Chiseling" and "Simultaneous Chiseling and Peeling" phases to separate the sheet. The lower panels provide corresponding visual snapshots for each phase across three synchronized perspectives.}
    \label{fig:timing}
\end{figure*}

We collected a total of 51 expert demonstrations, distributed across three training geometries: 17 on Square, 17 on Circle, and 17 on Octagon (see Fig. \ref{fig:5_sheets}). The data captures the five distinct phases of the task, as visualized in Fig. \ref{fig:timing}: Approach, Chiseling (high-frequency oscillation), Gripper Approach, Simultaneous Chiseling and Peeling, and Separation.

\subsection{Evaluation Metrics}
\label{subsec:metrics}
Contact-rich manipulation requires separating episode termination from safe, high-quality execution. In our setting, a trial can complete the separation motion while still being unsafe (e.g., force spikes, gouging, repeated slips). We therefore report Episode Completion and Task Success separately. 

\begin{enumerate}
    \item \textbf{Episode Completion Rate (\%):}
    Percentage of trials that reach the terminal condition (sheet fully separated along the intended perimeter) without operator intervention or safety abort (joint limits reached or experiencing unsafe contact forces). Episode Completion does not imply safe execution.
    
    \item \textbf{Peak Chisel Force ($\|\mathbf{F}\|_{\max}$):} Maximum measured chisel force magnitude $\|\mathbf{F}\|$ (from the 3-axis force) during the chiseling.

    \item \textbf{Execution Quality (0--3):}
    An ordinal score computed as follows:
    \begin{itemize}
        \item \textbf{3 (Excellent):} Stable contact and clean separation. No visible substrate gouging; no gripper slip; and the chisel force magnitude stays below a conservative threshold ($\|\mathbf{F}\|_{\max} < 30$ N) with no abrupt spikes.
        \item \textbf{2 (Acceptable):} Completes separation with minor jerk or one brief slip event, but no visible damage and no large force spikes ($30 \le \|\mathbf{F}\|_{\max} < 40$ N).
        \item \textbf{1 (Unsafe):} Completes separation but exhibits unsafe behavior such as repeated slips/jerk, tool skidding off the edge, or large force spikes ($\|\mathbf{F}\|_{\max} \ge 40$ N). This is considered a failure in safety-critical contexts.
        \item \textbf{0 (Aborts):} Triggers a safety stop and requires manual intervention. This could be due to reaching joint limits or locking due to excessive force.
    \end{itemize}
    All Execution Quality scores were assigned by the same evaluator while reviewing synchronized multi-view video and force logs to ensure consistency across methods. Force thresholds are used as an objective measure to reduce ambiguity.

    \item \textbf{Task Success Rate (\%):}
    A stricter metric than Episode Completion. A trial is counted as Task Success iff: (i) Episode Completion is achieved, and (ii) Execution Quality $\ge 2$.
    By definition, Task Success $\le$ Episode Completion. The gap between them corresponds to unsafe completions.

    \item \textbf{Execution Speed ($\times$ human performance):}
    It is the ratio of the average human task completion time for a specific sheet geometry ($T_{\text{human}}$) to the robot's completion time for the current trial ($T_{\text{robot}}$): $\text{Execution Speed} = T_{\text{human}} / T_{\text{robot}}$.

\end{enumerate}

\subsection{Baselines and Ablations}
\label{subsec: baselines}
We compare DPA-FTG against four baselines to isolate the contributions of our design:

\begin{enumerate}
    \item \textbf{DP-Vanilla \cite{chi2023diffusion}:} A standard vision-based Diffusion Policy.
    \item \textbf{DP with Force Feedback \cite{shukla_force-conditioned_nodate}}: A force-conditioned diffusion policy. The model is conditioned on the same slow context as our planner (vision + force + proprioception), but it still executes the chunk open-loop between replans.
    \item \textbf{Reactive Diffusion Policy \cite{xue_reactive_2025}:} We implement the slow-fast structure proposed in RDP: a slow latent diffusion policy for high-level intent and a fast reactive policy that runs at 20 Hz using high-frequency force feedback. In our setup, we use the wrist force data as the tactile signal. This baseline compares the slow-fast diffusion formulation with our approach. We evaluate the fast reactive policy at 20 Hz to remain faithful to its original formulation \cite{xue_reactive_2025}. As the overall system architecture was designed and tuned for this specific operating frequency. Therefore, this comparison highlights the difference between RDP's 20 Hz reactive correction and our approach's 60 Hz trajectory generation.
    \item \textbf{Behavioral Cloning (BC) \cite{mandlekar2021matters}:} A non-generative baseline trained with supervised learning at 60 Hz. A single network maps the observation to joint-velocity commands, using the same visual encoder backbone as our method and the same instantaneous proprioception and force inputs. 
\end{enumerate}

\subsection{Performance Comparison}
All evaluations are conducted on real hardware with non-trivial reset times (workpiece reloading, magnet re-engagement, and safety checks), which inherently constrains the total number of trials per method. To address this limitation, we report complementary metrics (episode completion, execution speed, and execution quality), evaluate across two unseen geometries, and include architectural ablations and failure-mode analyses. Despite the constrained sample size, the performance margins between DPA-FTG and the baselines remain consistent across all experimental axes, strongly supported by our qualitative evaluations and failure-mode analyses.

We first compare our method against Reactive Diffusion Policy (RDP) across 15 controlled in-distribution trials (5 per training geometry: Square/Circle/Octagon). In this setting, both methods achieve 100\% Episode Completion. Importantly, Episode Completion can be high even when the execution is unsafe; Task Success captures safety by requiring Execution Quality $\ge 2$ (Sec.~\ref{subsec:metrics}).

Table~\ref{tab:hier_vs_rdp} reports per-trial results. While RDP completes an episode, it frequently produces high force spikes and unstable oscillations, yielding low Execution Quality (mean 1.33) and a substantially lower Task Success rate (5/15 trials achieve Execution Quality $\ge 2$). In contrast, DPA-FTG maintains stable contact (lower peak force) and achieves consistently high Execution Quality. As detailed in Table \ref{tab:hier_vs_rdp}, RDP achieves a slightly higher execution speed compared to DPA-FTG. However, this speed advantage comes at the cost of safety.

\begin{table}[ht]
\centering
\caption{\textbf{Performance comparison on 15 controlled trials (5 per training geometry: Square/Circle/Octagon).} Both methods achieve 100\% Episode Completion. Peak $\|\mathbf{F}\|_{\max}$ is the maximum chisel force magnitude during the chiseling and simultaneous phases.}
\label{tab:hier_vs_rdp}
\setlength{\tabcolsep}{3.2pt}
\renewcommand{\arraystretch}{1.02}
\begin{tabular*}{\columnwidth}{@{\extracolsep{\fill}}lccc|ccc@{}}
\toprule
& \multicolumn{3}{c|}{\textbf{DPA-FTG (Ours)}} & \multicolumn{3}{c}{\textbf{RDP}} \\
\cmidrule(lr){2-4} \cmidrule(lr){5-7}
\textbf{Trial} &
\makecell{\textbf{Exec. Speed}\\($\times$human)} &
\makecell{\textbf{Exec. Qual.}\\(0--3)} &
\makecell{\textbf{Peak}\\$\|\mathbf{F}\|_{\max}$ (N)} &
\makecell{\textbf{Exec. Speed}\\($\times$human)} &
\makecell{\textbf{Exec. Qual.}\\(0--3)} &
\makecell{\textbf{Peak}\\$\|\mathbf{F}\|_{\max}$ (N)} \\
\midrule
Sq-1 & 0.86 & 3 & 26 & 0.90 & 1 & 50 \\
Sq-2 & 0.84 & 3 & 25 & 0.92 & 1 & 52 \\
Sq-3 & 0.87 & 3 & 24 & 0.89 & 1 & 47 \\
Sq-4 & 0.85 & 2 & 31 & 0.91 & 2 & 38 \\
Sq-5 & 0.83 & 3 & 27 & 0.86 & 2 & 36 \\
\addlinespace[2pt]
Ci-1 & 0.82 & 3 & 28 & 0.88 & 1 & 49 \\
Ci-2 & 0.84 & 3 & 26 & 0.90 & 1 & 45 \\
Ci-3 & 0.86 & 2 & 30 & 0.93 & 1 & 53 \\
Ci-4 & 0.88 & 3 & 27 & 0.91 & 1 & 51 \\
Ci-5 & 0.83 & 3 & 29 & 0.87 & 2 & 37 \\
\addlinespace[2pt]
Oc-1 & 0.85 & 3 & 25 & 0.89 & 1 & 46 \\
Oc-2 & 0.87 & 2 & 32 & 0.85 & 1 & 44 \\
Oc-3 & 0.89 & 3 & 26 & 0.88 & 2 & 39 \\
Oc-4 & 0.86 & 2 & 33 & 0.92 & 1 & 48 \\
Oc-5 & 0.84 & 3 & 27 & 0.84 & 2 & 35 \\
\midrule
\textbf{Mean} & \textbf{0.85} & \textbf{2.73} & \textbf{27.7} &
\textbf{0.89} & \textbf{1.33} & \textbf{44.7} \\
\bottomrule
\end{tabular*}
\end{table}

\subsection{Overall Performance Results}
Table~\ref{tab:method_comparison} summarizes performance across 15 in-distribution trials per method (training geometries, balanced across Square/Circle/Octagon). 

\begin{table}[ht]
\centering
\caption{
\textbf{Overall in-distribution performance metrics (15 trials per method).} 
Episode Completion denotes reaching terminal separation without triggering a safety abort. 
Task Success is a stricter metric, requiring full Episode Completion combined with an Execution Quality score $\ge 2$. 
Mean Execution Quality is computed over all trials, with aborted runs penalized as 0.
Execution Speed and Peak Force values are averaged over completed trials only. 
Standard Behavioral Cloning (BC) is marked as N/A due to consistent safety violations preventing valid data collection.
}

\label{tab:method_comparison}
\setlength{\tabcolsep}{4.0pt}
\renewcommand{\arraystretch}{1.05}
\begin{tabular*}{\columnwidth}{@{\extracolsep{\fill}}lccccc@{}}
\toprule
\textbf{Method} &
\makecell{\textbf{Episode Comp.}\\\textbf{Rate} (\%)} &
\makecell{\textbf{Task Success}\\\textbf{Rate} (\%)} &
\makecell{\textbf{Mean Exec.}\\\textbf{Quality} (0--3)} &
\makecell{\textbf{Mean Exec.}\\\textbf{Speed}} &
\makecell{\textbf{Mean Peak}\\$\|\mathbf{F}\|_{\max}$ (N)} \\
\midrule
DPA-FTG (Ours)          & 100.0 & 100.0 & 2.73 & 0.85 & 27.7 \\
RDP                     & 100.0 & 33.3 & 1.33 & 0.89 & 44.7 \\
DP w/ Force             & 86.7  & 20.0 & 1.13 & 0.79 & 46.0 \\
DP-Vanilla              & 66.7  & 13.3 & 0.80 & 0.72 & 50.5 \\
BC                      & NA  & NA  & NA & NA & NA \\
\bottomrule
\end{tabular*}
\end{table}

Our full DPA-FTG system achieves 100\% Episode Completion and the highest Task Success, indicating that the hierarchical planner with fast force-reactive decoder yields executions that are not only complete but also safe (high Execution Quality).

RDP achieves high Episode Completion (100\%) but substantially lower Task Success (33.3 \% or 5/15 trials). This is because many trials complete the execution but are unsafe due to force spikes (Execution Quality 1). DP-Vanilla performs poorly in this contact-rich setting. Even when it completes a run, open-loop action chunks cannot respond to fast force transients, resulting in large force overshoots and low Execution Quality. Adding force as conditioning (DP w/ Force) improves Episode Completion, but remains limited by open-loop execution within each chunk. BC was attempted using the same observation and action spaces, but it consistently triggered safety stops during preliminary rollouts. Therefore, we report it as NA in evaluations.

\subsection{Ablation Study}
We ablate key design choices within DPA-FTG (see Table \ref{tab:ablations}) to identify which components drive performance. All ablations are evaluated on the same 15 trials in-distribution protocol as Table \ref{tab:method_comparison}.

\begin{table}[ht]
\centering
\caption{
\textbf{Ablations of DPA-FTG (15 in-distribution trials; 5 per training geometry).}
The full system uses 
(i) diffusion-based task selection, 
(ii) VQ-VAE discrete skill quantization,
(iii) fast-loop force feedback, and 
(iv) decoder weight transfer.
Metrics and aggregation follow Table \ref{tab:method_comparison} (Execution Quality averages include aborts scored as 0; Execution Speed and Peak Force are averaged over completed trials).
}

\label{tab:ablations}
\setlength{\tabcolsep}{4.0pt}
\renewcommand{\arraystretch}{1.05}
\begin{tabular*}{\columnwidth}{@{\extracolsep{\fill}}lccccc@{}}
\toprule
\textbf{Variant} &
\makecell{\textbf{Episode Comp.}\\\textbf{Rate} (\%)} &
\makecell{\textbf{Task Success}\\\textbf{Rate} (\%)} &
\makecell{\textbf{Mean Exec.}\\\textbf{Qual.} (0--3)} &
\makecell{\textbf{Mean Exec.}\\\textbf{Speed}} &
\makecell{\textbf{Mean Peak}\\$\|\mathbf{F}\|_{\max}$ (N)} \\
\midrule
Full DPA-FTG (Ours) & 100.0 & 100.0 & 2.73 & 0.85 & 27.7 \\
\midrule
\makecell[l]{w/o Task Selection\\(fixed skill token)} & 100.0 & 53.3 & 1.67 & 0.87 & 41.2 \\
\hline
\makecell[l]{w/o Fast-Loop Force\\(no $F_t$ in $\pi_{\text{motion}}$)} & 100.0 & 73.3 & 2.13 & 0.86 & 34.5 \\
\hline
\makecell[l]{w/o VQ Quantization\\(continuous latent \\ to controller)} & 93.3 & 60.0 & 1.80 & 0.83 & 38.7 \\
\hline
\makecell[l]{w/o Weight Transfer\\(train $\pi_{\text{motion}}$ \\ from scratch)} & 86.7 & 46.7 & 1.47 & 0.80 & 42.5 \\
\bottomrule
\end{tabular*}
\end{table}

Removing task selection (fixed skill) preserves Episode Completion but degrades Task Success substantially, indicating that high-level context-dependent switching is essential for safe operation.
Removing fast-loop force feedback reduces Execution Quality and increases peak forces, confirming that high-frequency force closure is the primary mechanism preventing unstable oscillations during fracture events.
Removing VQ quantization increases failure, consistent with the role of the discrete codebook in preventing invalid mode-averaging in the latent space.
Finally, removing decoder weight transfer harms both Episode Completion and Execution Quality, showing that transferring kinematic structure from the VQ-VAE decoder materially improves data-efficiency and stability.

\subsection{Generalization to Novel Geometries}
To evaluate zero-shot geometric generalization, models are trained exclusively on Square, Circle, and Octagon sheets, then evaluated on two unseen shapes: (i) Hexagon (interpolation, $120^{\circ}$ corners) and (ii) Triangle (extrapolation, $60^{\circ}$ corners). Each method is run for 5 trials per unseen shape.

\begin{table}[t]
\centering
\caption{\textbf{Zero-shot generalization on unseen geometries (10 total trials; 5 trials/shape).} Metrics and aggregation follow Table~\ref{tab:method_comparison}.}
\label{tab:generalization}
\setlength{\tabcolsep}{3.5pt}
\renewcommand{\arraystretch}{1.05}
\begin{tabular*}{\columnwidth}{@{\extracolsep{\fill}}lccc|ccc@{}}
\toprule
& \multicolumn{3}{c|}{\textbf{Hexagon (unseen)}} & \multicolumn{3}{c}{\textbf{Triangle (unseen)}} \\
\cmidrule(lr){2-4} \cmidrule(lr){5-7}
\textbf{Method} &
\makecell{\textbf{Episode}\\\textbf{Comp.}\\\textbf{Rate} (\%) } &
\makecell{\textbf{Task}\\\textbf{Success}\\\textbf{Rate}(\%)} &
\makecell{\textbf{Mean}\\\textbf{Exec.}\\\textbf{Qual.}} &
\makecell{\textbf{Episode}\\\textbf{Comp.}\\\textbf{Rate} (\%) } &
\makecell{\textbf{Task}\\\textbf{Success}\\\textbf{Rate}(\%)} &
\makecell{\textbf{Mean}\\\textbf{Exec.}\\\textbf{Qual.}} \\
\midrule
DPA-FTG (Ours)          & 100 & 80 & 2.40 & 80 & 60 & 1.80 \\
RDP                     & 80  & 20  & 1.00 & 60  & 0   & 0.60 \\
DP w/ Force             & 60  & 20  & 0.80 & 40  & 0   & 0.40 \\
DP-Vanilla              & 40  & 0   & 0.40 & 20  & 0   & 0.20 \\
BC                      & NA  & NA  & NA   & NA  & NA  & NA \\
\bottomrule
\end{tabular*}
\end{table}

As shown in Table~\ref{tab:generalization}, DPA-FTG generalizes to both unseen geometries but performance decreases due to distribution shift. On the Hexagon, it achieves 80\% Task Success (4/5). On the Triangle, Task Success drops to 60\% (3/5) with 80\% Episode Completion (4/5). Failures on Triangle are dominated by brief tool slip near acute corners and early aborts, rather than failure to initiate contact.

In contrast, DP-Vanilla and DP w/ Force degrade sharply on Triangle due to open-loop commitment within action chunks, which frequently overshoots acute corners. RDP maintains high Episode Completion but low Task Success: it often finishes the separation while producing unsafe force spikes during corner transitions, lowering Execution Quality and, therefore, Task Success.

\subsection{Failure Mode Analysis}
\label{subsec:failure_modes}

While DPA-FTG achieved a high success rate on the zero-shot generalization set, it encountered 3 failures out of 10 trials: 2 unsafe completions and 1 abort (Table~\ref{tab:failure_outcomes}). 

\begin{table}[ht]
\centering
\caption{\textbf{Outcome decomposition on the zero-shot evaluation set from Table~\ref{tab:generalization}.}
The analysis focuses on the 10 trials involving unseen geometries (Hexagon and Triangle). Unsafe Completion denotes trials where the robot separated the sheet but exceeded force safety thresholds (Quality=1). Abort denotes trials terminated early by the safety system (Quality=0).}
\label{tab:failure_outcomes}
\setlength{\tabcolsep}{4.0pt}
\renewcommand{\arraystretch}{1.05}
\begin{tabular*}{\columnwidth}{@{\extracolsep{\fill}}lccc@{}}
\toprule
\textbf{Method} & \textbf{Success} & \textbf{Unsafe Completion} & \textbf{Abort} \\
\midrule
DPA-FTG (Ours)                & 7/10 & 2/10 & 1/10 \\
RDP                           & 1/10 & 6/10 & 3/10 \\
DP w/ Force (open-loop)       & 1/10 & 4/10 & 5/10 \\
DP-Vanilla                    & 0/10 & 3/10 & 7/10 \\
\bottomrule
\end{tabular*}
\end{table}

\begin{figure*}[t]
    \centering
    \includegraphics[width=\linewidth]{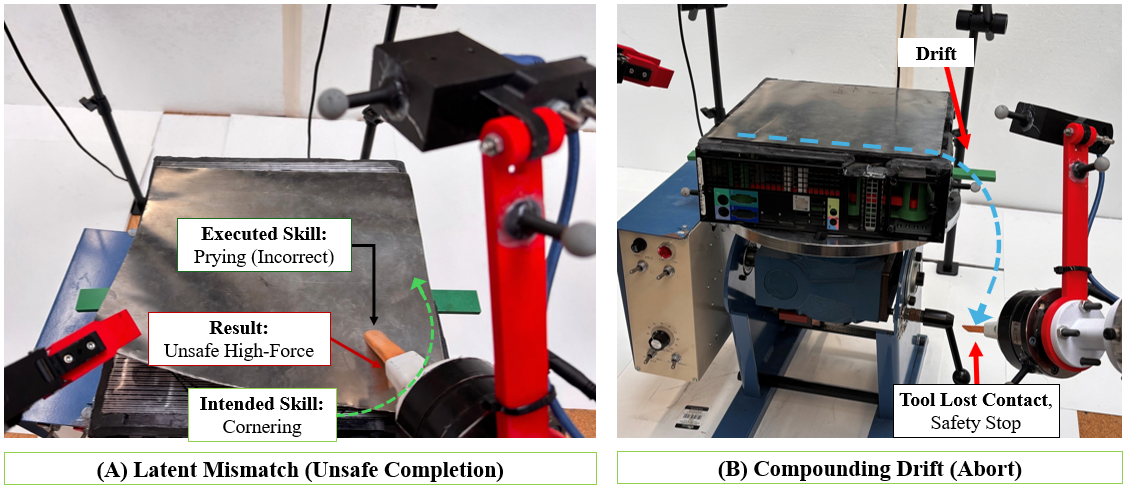} 
    \caption{\textbf{Failure Mode Analysis.} Illustration of two distinct failure modes during the task. (A) Latent Mismatch (Unsafe Completion): The system executes an incorrect skill, Prying, instead of the intended Cornering sequence. This results in an Unsafe High-Force pry action on the workpiece. (B) Compounding Drift (Abort): The tool drifts away from the intended trajectory (blue dashed line), causing it to lose contact with the surface. This deviation triggers a Safety Stop.}
    \label{fig:failure_modes}
\end{figure*}

To understand the mechanisms behind these failures, we identified two distinct failure modes when generalizing to novel geometries as shown in Fig. \ref{fig:failure_modes}.

\textbf{A. Latent Mismatch (Cause of Unsafe Completions)}
This mode accounted for the two unsafe completion trials. It occurs when the high-level planner misinterprets a geometric feature and selects an incorrect skill primitive. For example, at the sharp  corner of the Triangle, the planner occasionally generated a latent code  corresponding to a prying motion rather than the required cornering motion. Because the low-level controller is conditioned on this incorrect intent, it executes a valid motion that is inappropriate for the context. This results in a sudden, high-force interaction and drops the Execution Quality score to 1. There are two plausible reasons for this failure mode (i) limited demonstration coverage of acute-corner transitions, which reduces the probability the planner assigns to the correct cornering primitive under distribution shift, and (ii) limited geometric sensitivity of the ResNet-18 slow-context encoder.

\textbf{(B) Compounding Drift (Cause of Aborts).}
This mode accounted for the single aborted trial. Failures of this type originate from a loss of tool-edge contact, such as a lateral slip during the approach phase. Once the chisel tip leaves the adhesive interface, the state falls out of the distribution represented in the expert demonstrations. Lacking an explicit recovery skill in its library, the low-level policy fails to correct the pose error. Instead, the error accumulates over time until the tool jams against the substrate or drifts beyond the workspace limits, triggering an automatic safety stop to prevent damage.

\vspace{-1em}
\section{Conclusions}
\vspace{-1em}
\label{sec:conclusion}
\noindent In this paper, we addressed a critical limitation of diffusion policies that has hindered their application to dynamic, contact-rich manipulation tasks: their inability to perform high-frequency, reactive control. We introduced the Diffusion Policy Augmented by Fast Trajectory Generation (DPA-FTG), an architecture that resolves this issue by decoupling low-frequency strategic planning from high-frequency feedback control. Our approach uses a conditional diffusion model for multi-modal planning at a low frequency, and a dedicated, lightweight learned controller provides real-time, force-sensitive adjustments at a high frequency.

Our experimental evaluation on a challenging bimanual sheet separation task demonstrated the benefits of this hierarchical design. The full DPA-FTG system significantly outperformed baselines that lacked either high-frequency reactive control or high-level strategic planning, achieving a high task completion rate and stable motions. These results validate our claim that a hierarchical architecture is an effective solution for extending the capabilities of diffusion policies into the domain of high-frequency, forceful manipulation.

While our approach proved effective, it has some limitations:
\begin{itemize}
    \item The zero-shot generalization experiments highlighted challenges in extrapolating to unseen acute geometries, which resulted in the Latent Mismatch failure mode (Section \ref{subsec:failure_modes}). This limitation fundamentally stems from a combination of constrained demonstration coverage and the geometric capacity of the current visual encoder. Consequently, a critical direction for future work is to improve the system's generalization. We plan to address this by enriching the training distribution with targeted acute-corner demonstrations and upgrading the ResNet-18 backbone to a more spatially aware visual architecture, thereby improving the planner's ability to select correct primitives under distribution shifts.

    \item The current framework relies on a learned latent primitive vocabulary that forms the interface between the two policy levels. A potential direction for future research is to develop methods for learning this vocabulary of primitives automatically from unlabeled demonstration data.

    \item Finally, we aim to integrate recent advances in sampling acceleration by replacing the high-level diffusion policy in our architecture with a one-step Consistency Policy. Such a system could further reduce planning latency by combining the benefits of a fast planner with a fast controller.
\end{itemize}


\bibliography{references}

\end{document}